\pdfoutput=1
\documentclass[11pt]{article}
\usepackage{emnlp2021}
\usepackage{times,latexsym}
\usepackage{url}
\usepackage{hyperref}
\usepackage[T1]{fontenc}
\usepackage{times}
\usepackage{subcaption}
\usepackage{graphicx}
\usepackage{longtable}
\usepackage{multicol}
\usepackage{multirow}
\usepackage{makecell}
\usepackage{placeins}
\usepackage{varioref}
\usepackage{algorithm}
\usepackage{microtype}
\usepackage{natbib}
\usepackage{bm}

\usepackage{latexsym}
\usepackage{amsmath}
\usepackage{float} 
\usepackage{amssymb}

\usepackage{amsthm}

\theoremstyle{definition}

\usepackage{algorithmicx}
\usepackage{algpseudocode}
\usepackage{changes}

\usepackage{xspace,mfirstuc,tabulary}

\newif\iftaclinstructions
\taclinstructionsfalse 
\iftaclinstructions
\renewcommand{\confidential}{}
\renewcommand{\anonsubtext}{}
\newcommand{\instr}
\fi
\usepackage{tablefootnote}
\title{Code-switched inspired losses for generic spoken dialog representations}

\author{\textbf{ Emile Chapuis\textsuperscript{\rm 1}}\footnote[1]{Equal contribution} , \textbf{Pierre Colombo}\textsuperscript{\rm 1,2}\thanks{$\rm \text{ }$ stands for equal contribution} , 
 \\ \textbf{Matthieu Labeau\textsuperscript{\rm 1}, Chloe Clavel\textsuperscript{\rm 1}}\\
\textsuperscript{\rm 1}LTCI, Telecom Paris, Institut Polytechnique de Paris, \\ \textsuperscript{\rm 2}IBM GBS France \\
\textsuperscript{\rm 1}firstname.lastname@telecom-paris.fr, \\
\textsuperscript{\rm 2}pierre.colombo@ibm.com \\
}

\date{}

\begin{document}
\maketitle
\begin{abstract}
Spoken dialog systems need to be able to handle both multiple languages and multilinguality inside a conversation (\textit{e.g} in case of code-switching). In this work, we introduce new pretraining losses tailored to learn multilingual spoken dialog representations. The goal of these losses is to expose the model to code-switched language. To scale up training, we automatically build a pretraining corpus composed of multilingual conversations in five different languages (French, Italian, English, German and Spanish) from \texttt{OpenSubtitles}, a huge multilingual corpus composed of 24.3G tokens. We test the generic representations on \texttt{MIAM}, a new benchmark composed of five dialog act corpora on the same aforementioned languages as well as on two novel multilingual downstream tasks (\textit{i.e} multilingual mask utterance retrieval and multilingual inconsistency identification). Our experiments show that our new code switched-inspired losses achieve a better performance in both monolingual and multilingual settings.


\end{abstract}

\section{Introduction} 
A crucial step in conversational AI is the identification of underlying information of the user's utterance (\textit{e.g} communicative intent or dialog acts, and emotions). This requires modeling utterance-level information \cite{corner2,corner1}, to capture immediate nuances of the user utterance; and discourse-level features \cite{context}, to capture patterns over long ranges of the conversation. An added difficulty to this modeling problem is that most people in the world are bilingual \cite{grosjean2013psycholinguistics}: therefore, progress on these systems is limited by their inability to process more than one language (English being the most frequent). For example, many people use English as a ``workplace" language but seamlessly switch to their native language when the conditions are favorable \cite{heredia2001bilingual}. Thus, there is a growing need for understanding dialogs in a multilingual fashion \cite{ipsic1999multilingual,joshi2020state,ruder2019survey}. Additionally, when speakers share more than one language, they inevitably will engage in code-switching \cite{sankoff1981formal,gumperz1982discourse,milroy1995one,auer2013code,parekh2020understanding}: switching between two different languages. Thus, spoken dialog systems need to be cross lingual (\textit{i.e} able to handle different languages) but also need to model multilinguality inside a conversation \cite{ahn2020code}.

In this paper, we focus on building generic representations for dialog systems that satisfy the aforementioned requirements. Generic representations have led to strong improvements on numerous natural language understanding tasks, and can be fine-tuned when only small labelled datasets are available for the desired downstream task \cite{mikolov2013exploiting,bert,albert,roberta,xlnet}. While there has been a growing interest in pretraining for dialog \cite{mehri2019pretraining,zhang2019dialogpt}, the focus has mainly been on English datasets. Thus, these works can not be directly applied to our multilingual setting.
Additionally, available multilingual pretraining objectives \cite{lample2019cross,liu2020multilingual,xue2020mt5,qi2021prophetnet} face two main limitations when applied to dialog modeling: (1) they are a generalization of monolingual objectives that use flat input text, whereas hierarchy has been shown to be a powerful prior for dialog modeling. This is a reflection of a dialog itself, for example, context plays an essential role in the labeling of dialog acts. (2) The pretraining objectives are applied separately to each language considered, which does not expose the (possible) multilinguality inside a conversation (as it is the case for code-switching)~\cite{winata2021multilingual}\footnote{We refer to code-switching at the utterance level, although it is more commonly studied at the word or span level \cite{poplack1980sometimes,banerjee2018dataset,bawa2020multilingual,fairchild2017determiner}}.


Our main contributions are as follows:
\\1. \textit{We introduce a set of code-switched inspired losses as well as a new method to automatically obtain several million of conversations with multilingual input context in different languages}. There has been limited work on proposing corpora with a sufficient amount of conversations that have multilingual input context. Most of this work focuses on social media, or on corpora of limited size. Hence, to test our new losses and scale up our pretraining, we automatically build a pretraining corpus of multilingual conversations, each of which comprises several languages, by leveraging the alignments available in OpenSubtitles (\texttt{OPS}). 
\\2. \textit{We showcase the relevance of the aforementioned losses and demonstrate that it leads to better performances on downstream tasks, that involve both monolingual conversations and multilingual input conversations.} For monolingual evaluation, we introduce the \textbf{M}ultilingual d\textbf{I}alog\textbf{A}ct bench\textbf{M}ark (\texttt{MIAM}): composed of five datasets in five different languages annotated with dialog acts. Following \citet{mehri2019pretraining,lowe2016evaluation}, we complete this task with both contextual inconsistency detection and next utterance retrieval in these five languages. For multilingual evaluation, due to the lack of code-switching corpora for spoken dialog, we create two new tasks: contextual inconsistency detection and next utterance retrieval with multilingual input context. The datasets used for these tasks are unseen during training and automatically built from \texttt{OPS}.
\\In this work, we follow the recent trend~\cite{albert,tiny} in the NLP community that aims at using models of limited size that can both be pretrained with limited computational power and achieve good performance on multiple downstream tasks. The languages we choose to work on are English, Spanish, German, French and Italian.\footnote{Although our pretraining can be easily generalised to 62 languages, we use a limited number of languages to avoid exposure to the so-called ``curse of multilinguality'' \cite{conneau2019unsupervised}}. \texttt{MIAM} is available in Datasets \cite{2020HuggingFace-datasets} \url{https://huggingface.co/datasets/miam}.


\section{Model and training objectives} 
\textbf{Notations} We start by introducing the notations. We have a set $D$ of contexts (\textit{i.e} truncated conversations), i.e., $D = (C_1,C_2,\dots,C_{|D|})$. Each context $C_i$ is composed of utterances $u$, i.e $C_i= (u^{L_1}_1,u^{L_2}_2,\dots,u^{L_{|C_i|}}_{|C_i|})$ where $L_i$ is the language of utterance $u_i$\footnote{In practice, we follow \cite{sankar2019neural} and set the context length to $5$ consecutive utterances.}. At the lowest level, each utterance $u_i$ can be seen as a sequence of tokens, i.e $u^{L_i}_i = (\omega^i_1, \omega^i_2, \dots, \omega^i_{|u_i|})$. For \texttt{DA} classification $y_i$ is the unique dialog act tag associated to $u_i$. In our setting, we work with a shared vocabulary $\mathcal{V}$ thus $\omega^i_j \in \mathcal{V}$ and  $\mathcal{V}$ is language independent.

\subsection{Related work}\label{ssec:multi_pretraining}
\noindent\textbf{Multilingual pretraining}. Over the last few years, there has been a move towards pretraining objectives, allowing models to produce general multilingual representations that are useful for many tasks. However, they focus on the word level \cite{gouws2015bilbowa,mikolov2013exploiting,faruqui2014improving} or the utterance level \cite{bert,lample2019cross,eriguchi2018zero}. \citet{winata2021multilingual} shows that these models obtain poor performances in presence of code-switched data.
\\\noindent\textbf{Pretraining to learn dialog representation}. Current research efforts made towards learning dialog representation are mainly limited to the English language \cite{henderson2019convert,mehri2019pretraining,chapuis2020hierarchical} and introduce objectives at the dialog level such as next-utterance retrieval, next-utterance generation, masked-utterance retrieval, inconsistency identification or generalisation of the cloze task \cite{cloze}. To the best of our knowledge, this is the first work to pretrain representations for spoken dialog in a multilingual setting. 
\\\noindent\textbf{Hierarchical pretraining} As we are interested in capturing information at different granularities, we follow the hierarchical approach of \citet{chapuis2020hierarchical} and decompose the pretraining objective in two terms: the first one for the utterance level and the second one to capture discourse level dependencies. Formally, the global hierarchical loss can be expressed as:
\begin{equation}\label{eq:h_loss}
    \mathcal{L}(\theta) = \underbrace{\lambda_u \times \mathcal{L}^u(\theta)}_{\text{utterance level}} +   \underbrace{\lambda_d \times \mathcal{L}^d(\theta)}_{\text{dialog level}}.
\end{equation}
These losses rely on a hierarchical encoder~\cite{crf_chen, sota_swda_1} composed of two functions $f^u$ and $f^d$:
\begin{align}\label{eq:multi_obj}
        \mathcal{E}_{u^{L_i}_i} &= f^u_\theta(\omega^i_1, \hdots, \omega^i_{|u_i|}), \\
        \mathcal{E}_{C_j} &= f^d_\theta( \mathcal{E}_{u^{L_1}_1}, \hdots, \mathcal{E}_{u^{L_{|C_j|}}_{|C_j|}}),
\end{align}
where $\mathcal{E}_{u^{L_i}_i} \in \mathbb{R}^{d_u}$ is the embedding of $u^{L_i}_i$ and $\mathcal{E}_{C_j} \in \mathbb{R}^{d_d}$ the embedding of $C_j$. The encoder is built on transformer layers.
\subsection{Utterance level pretraining}\label{ssec:Pretraining}
To train the first level of hierarchy (\textit{i.e} $f^u_\theta$), we use a Masked Utterance Modelling (MUM) loss \cite{bert}. Let $u^{L_i}_i$ be an input utterance and $\Tilde{u}^{L_i}_i$ its corrupted version, obtained after masking a proportion $p_\omega$ of tokens, the set of masked indices is denoted $\mathcal{M}_{\omega}$. The set of masked tokens is denoted $\Omega$.
The probability of the masked token given $\Tilde{u}^{L_i}_i$ is given by:
\begin{equation}
    p(\Omega |\Tilde{u}^{L_i}_i)= \prod_{t \in \mathcal{M}_{\omega}} p_{\theta}(\omega^i_t | \Tilde{u}^{L_i}_i).
     \label{eq:mlm_loss}
\end{equation} 
\subsection{Dialog level pretraining}
The goal of the dialog level pretraining is to ensure that the model learns dialog level dependencies (through $f^d_\theta$), \textit{i.e} the ability to handle multi-lingual input context.
\\\textbf{Generic framework} Given ${C}_k$ an input context, a proportion $p_{\mathcal{C}}$ of utterances is masked to obtained the corrupted version $\Tilde{C}_k$. The set of masked utterances is denoted $\mathcal{U}$ and the set of corresponding masked indices $\mathcal{M}_{u}$. The probability of $\mathcal{U}$ given $\Tilde{C}_k$ is:
\begin{equation}
    p(\mathcal{U} |\Tilde{C}_k)= \prod_{t \in \mathcal{M}_{\mathcal{U}}}\prod_{j=0}^{|u_t|-1} p_{\theta}(\omega^t_j | \omega^t_{1:j-1} , \Tilde{C}_k).
     \label{eq:mlm_loss}
\end{equation}
As shown in \autoref{eq:mlm_loss}, a masked sequence is predicted one word per step. As an example, at the j-th step, the prediction of $\omega^t_j$ is made given $(\omega^t_{1:j-1} , \Tilde{C}_k)$ where $\omega^t_{1:j-1} =(\omega^t_{1},\cdots,\omega^t_{1:j-1})$. In the following, we describe different procedures to build $\mathcal{M}_{\mathcal{U}}$ and $\Tilde{C}_k$ used in \autoref{eq:mlm_loss}.
\begin{table*}[]
    \centering
    \resizebox{\textwidth}{!}{\begin{tabular}{c|c|l|l}\hline
   Index & Speaker & Monolingual  Input &  Multilingual  Input \\\hline
 0 & A        & Good afternoon.  & Good afternoon.  \\
 1 & A &I'm here to see Assistant Director Harold Cooper. &Je suis ici pour voir l'assistant directeurr Harold Cooper.  \\
 2 & B &Do you have an appointment? & Do you have an appointment?  \\
 3 & A&I do not. & Non. \\
 4 & A &Tell him it's Raymond Reddington. & Dites lui que c'est Raymond Reddington.
 \\\hline
    \end{tabular}}
    \caption{Example of automatically built input context from \texttt{OPS}.}
    \label{tab:example_multinlingual_monolingual}
\end{table*}

\subsubsection{Masked utterance generation (MUG)} 
The MUG loss aims at predicting the masked utterance from a monolingual input context. As the vocabulary is shared, this loss will improve the alignment of conversations at the dialog level. This loss ensures that the model will be able to handle monolingual conversations in different languages.
\\\textbf{Training Loss} We rely on \autoref{eq:mlm_loss} for MUG. The input context is composed of utterances in the same language, \textit{i.e} $\forall k,C_k= (u_1^{L_k},\cdots,u_{|C_k|}^{L_k})$. The mask is randomly chosen among all the positions.
\\\textbf{Example} Given the monolingual input context given in \autoref{tab:example_multinlingual_monolingual}, a random mask (\textit{e.g} $[0,3]$) is chosen among the positions $[0,1,2,3,4]$. The masked utterances are replaced by \texttt{[MASK]} tokens to obtain $\Tilde{C}_k$ and a decoder attempts to generate them.

\subsubsection{Translation masked utterance generation (TMUG)}
The previous objectives are self-supervised and cannot be employed with parallel data when available. In addition, these losses do not expose the model to multilinguality inside the conversation. The TMUG loss addresses this limitation using a translation mechanism: the model learns to translate the masked utterance in a new language.
\\\textbf{Training Loss} We use \autoref{eq:mlm_loss} for TMUG with a bilingual input context $C_k$. $C_k$ contains two different languages (\textit{i.e} $L$ and $L^\prime$) $\forall k,C_k= (u_1^{L_1},\cdots,u_{|C_k|}^{L_k})$ with $L_i \in \{L,L^\prime\}$. The masked positions $\mathcal{M}_u$ are all the utterances in language $L^\prime$. Thus $\tilde{C}_k$ is a monolingual context.
\\\textbf{Example} Given the multilingual input context given in \autoref{tab:example_multinlingual_monolingual}, the positions $[3,4]$ are masked with sequences of \texttt{[MASK]} and the decoder will generate them in French. See \autoref{ssec:additionnal_pretriained} for more details on the generative pretraining.

\subsubsection{Multilingual masked utterance generation (MMUG)}
In the previous objectives, the model is exposed to monolingual input only. MMUG aims at relaxing this constraint by considering multilingual input context and generating the set of masked utterances in any possible target language. 
\\\textbf{Training Loss} Given a multi-lingual input context $C_k=(u_1^{L_1},\cdots,u_{|C_k|}^{L_{|C_k|}})$. A random set of indexes is chosen and the associated utterances are masked. The goal remains to generate the masked utterances. 
\\\textbf{Example} In \autoref{tab:example_multinlingual_monolingual}, the positions $[2,3]$ are randomly selected from the available positions $[0,1,2,3,4]$. Given these masked utterances the model will generate $2$ in Italian and $3$ in Spanish. 
MMUG is closely related to code-switching as it exposes the model to multilingual context and the generation can be carried out in any language.

\subsection{Pretraining corpora}\label{ssec:pretraining_corpore}
There is no large corpora freely available that contains a large number of transcripts of well segmented multilingual spoken conversation\footnote{Specific phenomena appear (\textit{e.g.} disfluencies \cite{dinkar2018disfluencies}, filler words \cite{dinkar2020importance}) when working with spoken language, as opposed to written text.} with code switching phenomenon. Collecting our pretraining corpus involves two steps: the first step consists of segmenting the corpus into conversations, in the second step, we obtain aligned conversations.
\\\noindent\textbf{Conversation segmentation} Ideal pretraining corpora should contain multilingual spoken language with dialog structure. In our work, we focus on \texttt{OPS} \cite{opensub}\footnote{http://opus.nlpl.eu/OpenSubtitles-alt-v2018.php} because it is the only free multilingual dialog corpus ($62$ different languages). After preprocessing, \texttt{OPS} contains around $50M$ of conversations and approximately $8$ billion of words from the five different languages (\textit{i.e} English, Spanish, German, French and Italian). \autoref{tab:monlingual_open} gathers statistics on the considered multilingual version of \texttt{OPS}. To obtain conversations from \texttt{OPS}, we consider that two consecutive utterances are part of the same conversation if the inter-pausal unit \cite{koiso1998analysis} (\textit{i.e} silence between them) is shorter than $\delta_T=6s$. If a conversation is shorter than the context size $T$, they are dropped and utterance are trimmed to $50$ (for justification see \autoref{fig:utt_hist}).
\\\noindent\textbf{Obtaining aligned conversations} We take advantage of the alignment files provided in \texttt{OPS}. They provide an alignment between utterances written in two different languages. It allows us to build aligned conversations with limited noise (solely high confidence alignments are kept). Statistics concerning the aligned conversations can be found in \autoref{tab:multi_open} and an example of automatically aligned context can be found in \autoref{tab:example_multinlingual_monolingual}. The use of more advanced methods to obtain more fine-grained alignment (\textit{e.g} word level alignment, span alignment inside an utterance) is left as future work.
\begin{table}[]
    \centering
   \resizebox{0.45\textwidth}{!}{\begin{tabular}{c|ccccc}\hline
                 & de & en & es  & fr & it  \\\hline
\# movies   &46.5K&446.5K&234.4K&127.2K&134.7K \\
\# conversations  & 1.8M&18.2M&10.0M&5.2M&4.2M \\
\# tokens  & 363.6M&3.7G&1.9G&1.0G&994.7M \\
    \hline\end{tabular}}
    \caption{Statistics of the processed version of \texttt{OPS}.} 
    \label{tab:monlingual_open}
\end{table}
\begin{table}[]
    \centering
    \resizebox{0.45\textwidth}{!}{ \begin{tabular}{c|ccccc}\hline
                 & \multicolumn{1}{c}{de-en} & \multicolumn{1}{c}{de-es} & \multicolumn{1}{c}{de-fr} & \multicolumn{1}{c}{de-it} & \multicolumn{1}{c}{en-es}  \\\hline
\# utt. &23.4M&19.9M&17.1M&14.1M&63.5M \\
\# tokens. &217.3M&194.1M&167.0M&139.5M&590.9M \\\hline
                &    \multicolumn{1}{c}{en-fr} & \multicolumn{1}{c}{en-it} & \multicolumn{1}{c}{es-fr} & \multicolumn{1}{c}{es-it} & \multicolumn{1}{c}{fr-it} \\\hline
\# utt.  &44.2M&36.7M&37.9M&31.4M&23.8M \\
\# tokens. &413.7M&347.1M&362.1M&304.6M&248.5M \\\hline
    \end{tabular}}
    \caption{Statistics of the processed version of the alignment files from \texttt{OPS}.}
    \label{tab:multi_open}
\end{table}


\section{Evaluation framework}
This section presents our evaluation protocol. It involves two different types of evaluation depending on the input context. The first group of experiences consists in multilingual evaluations with monolingual input context and follows classical downstream tasks \cite{finch2020towards,dziri2019evaluating} including sequence labeling \cite{colombo2020guiding}, utterance retrieval \cite{mehri2019pretraining} or inconsistency detection. The second group focuses on multilingual evaluations with multilingual context.
\subsection{Dialog representations evaluation}
\subsubsection{Monolingual context}\label{ssec:evaluating_multimodal}
\textbf{Sequence labeling tasks}. The ability to efficiently detect and model discourse structure is an important step toward modeling spontaneous conversations. A useful first level of analysis involves the identification of dialog act (\texttt{DA}) \cite{hmm_dialog} thus \texttt{DA} tagging is commonly used to evaluate dialog representations. However, due to the difficulty to gather language-specific labelled datasets, multilingual sequence labeling such as \texttt{DA} labeling remains overlooked.
\\\textbf{Next-utterance retrieval (\texttt{NUR})} The utterance retrieval task \cite{duplessis2017utterance,saraclar2004lattice} focuses on evaluating the ability of an encoder to model contextual dependencies. \citet{lowe2016evaluation} suggests that NUR is a good indicator of how well context is modeled.
\\ \textbf{Inconsistency Identification (\texttt{II})} Inconsistency identification is the task of finding inconsistent utterances within a dialog context \cite{incosistency_dialog}. The perturbation is as follow: one utterance is randomly replaced, the model is trained to find the inconsistent utterance.\footnote{To ensure fair comparison, contrarily to \citet{mehri2019pretraining} the pretraining is different from the evaluation tasks.}

\subsubsection{Multilingual context}
To the best of our knowledge, we are the first to probe representation for multi-lingual spoken dialog with multilingual input context. As there is no labeled code-switching datasets for spoken dialog (research focuses on on synthetic data \cite{stymne2020evaluating}, social media \cite{pratapa2018word} or written text \cite{khanuja2020gluecos,tan2021code} rather than spoken dialog). Thus we introduce two new downstream tasks with automatically built datasets: Multilingual Next Utterance Retrieval (\texttt{mNUR}) and Multilingual Inconsistency Identification (\texttt{mII}). To best assess the quality of representations, for both \texttt{mII} and \texttt{mNUR} we choose to work with train/test/validation datasets of 5k conversations. The datasets, unseen during training, are built using the procedure described in \autoref{ssec:pretraining_corpore}.
\\\textbf{Multilingual next utterance retrieval}. \texttt{mNUR} consists of finding the most probable next utterance based on an input conversation. The evaluation dataset is built as follow: for each conversation in language $L$ composed of $T$ utterances, a proportion $p_{L^\prime}$ of utterances is replaced by utterances in language $L^\prime$. $D$ utterances that we call distractors\footnote{$D$ is set to $9$ according to \cite{ubuntu}} in language $L$ or $L^\prime$ from the same movie. For testing, we frame the task as a ranking problem and report the recall at $N$ (R@N)~\cite{schatzmann2005quantitative}.   
\\\textbf{Multilingual inconsistency identification}. The task of \texttt{mII} consists of identifying the index of the inconsistent sentences introduced in the conversation. Similarly to the previous task: for each conversation in language $L$ composed of $T$ utterances, a proportion $p_{L^\prime}$ is replaced by utterances in language $L^\prime$, a random index is sampled from $[1,T]$ and the corresponding utterance is replaced by a negative utterance taken from the same movie. 

\subsection{Multilingual dialog act benchmark}\label{ssec:da}
\texttt{DA}s are semantic labels associated with each utterance in a conversational dialog that indicate the speaker's intention (examples are provided in \autoref{tab:comp_ex}).
A plethora of freely available dialog act dataset \cite{datase_swda,dataset_mrda,dataset_dailydialog}) 
has been proposed to evaluate \texttt{DA} labeling systems in English. However, constituting a multilingual dialog act benchmark is challenging \cite{ribeiro2019multilingual}. We introduce \textbf{M}ultilingual d\textbf{I}alogue \textbf{A}ct bench\textbf{M}ark (in short \texttt{MIAM}). This benchmark gathers five free corpora that have been validated by the community, in five different European languages (\textit{i.e.} English, German, Italian, French and Spanish). We believe that this new benchmark is challenging as it requires the model to perform well along different evaluation axis and validates the cross-lingual generalization capacity of the representations across different annotation schemes and different sizes of corpora.
\\\textbf{\texttt{DA} for English} For English, we choose to work on the \texttt{MapTask} corpus. It consists of conversations where the goal of the first speaker is to reproduce a route drawn only on the second speaker’s map, with only vocal indications. We choose this corpus for its small size that will favor transfer learning approaches (27k utterances).
\\\textbf{\texttt{DA} for Spanish} Spanish research on \texttt{DA} recognition mainly focuses on three different datasets \texttt{Dihana}, CallHome Spanish \cite{post2013improved} and DIME \cite{coria2005predicting,olguin2006predicting}. \texttt{Dihana} is the only available corpora that contains free \texttt{DA}  annotation \cite{dihana_1}. It is a spontaneous speech corpora \cite{benedi2006design} composed of $900$ dialogs from $225$ users. Its acquisition was carried out using a Wizard of Oz setting \cite{fraser1991simulating}. For this dataset, we focus on the first level of labels which is dedicated to the task-independent \texttt{DA}.
\\\textbf{\texttt{DA} for German} For German, we rely on the VERBMOBIL (\texttt{VM2}) dataset \cite{kay1992verbmobil}. This dataset was collected in two phases: first, multiple dialogs were recorded in an appointment scheduling scenario, then each utterance was annotated with \texttt{DA} using $31$ domain-dependent labels. The three most common labels (\textit{i.e.} inform, suggest and feedback) are highly related to the planning nature of the data. 
\\\textbf{\texttt{DA} for French} Freely available to academic and nonprofit research datasets are limited in the french language as most available datasets are privately owned. We rely on the french dataset from the Loria Team \cite{barahona2012building} (\texttt{LORIA}) where the collected data consists of approximately $1250$ dialogs and $10454$ utterances. The tagset is composed of $31$ tags.
\\\textbf{\texttt{DA} for Italian} For Italian, we rely on the \texttt{Ilisten} corpora \cite{basile2018overview}. The corpus was collected in a Wizard of Oz setting and contains a total of $60$ dialogs transcripts, $1,576$ user dialog turns and $1,611$ system turns. The tag set is composed of $15$ tags.
\\\textbf{Metrics}: There is no consensus on the evaluation metric for \texttt{DA} labelling (e.g., \citet{weighted_preproc,weighted_no} use a weighted F-score while \citet{accuracy_fscore} report accuracy). We follow \citet{chapuis2020hierarchical} and report accuracy.

\subsection{Baseline encoders for downstream tasks}
The encoders that will serve as baselines can be divided into two different categories: hierarchical encoders based on GRU layers ($\mathcal{HR}$) and pretrained encoders based on Transformer cells \cite{attention_is}. The first group achieve SOTA results on several sequence labelling tasks~\cite{self_attention,sota_swda_1}. The second group can be further divided in two groups: language specific ($BERT$) and multilingual BERT ($mBERT$) \footnote{Details of language specific BERT and on baseline models can be found in \autoref{ssec:additionnal_pretriained} and in \autoref{ssec:baseline_details} respectively. } and pretrained hierarchical transformers from \cite{zhang2019hibert} ($\mathcal{HT}$) are used as a common architecture to test the various pretraining losses. 
\\\textbf{Tokenizer} We will work with both language specific and multlingual tokenizer. Model with multilingual tokenizer will be referred with a m (\textit{e.g} $mBERT$ as opposed to $BERT$).
    



\section{Numerical results}
In this section, we empirically demonstrate the effectiveness of our code-switched inspired pretraining on downstream tasks involving both monolingual and multilingual input context.

\subsection{Monolingual input context}
\subsubsection{\texttt{DA} labeling}\label{ssec:da_labelling}
\noindent\textbf{Global analysis.} \autoref{tab:multilingual_da} reports the results of the different models on \texttt{MIAM}. \autoref{tab:multilingual_da} is composed of two distinct groups of models: \textit{language specifics models} (with language-specific tokenizers) and  \textit{multilingual models} (with a multilingual tokenizer denoted with a $m$ before the model name). Overall, we observe that $mMUG$ augmented with both $TMUG$ and $MMUG$ gets a boost in performance ($1.8\%$ compared to $mMUG$ and $2.6\%$ compared to a mBERT model with a similar number of parameters). This result shows that the model benefits from being exposed to aligned bilingual conversations and that our proposed losses (\textit{i.e.} $TMUG$ and $MMUG$) are useful to help the model to better catch contextual information for \texttt{DA} labeling. 
\\\noindent\textbf{Language-specific v.s. multilingual models}. By comparing the performances of $\mathcal{HR}$ (with either a CRF or MLP decoder), we can notice that for these models on \texttt{DA} labelling it is better to use a multilingual tokenizer. As multilingual tokenizers are not tailored for a specific language and have roughly twice as many tokens than their language-specific counterparts, one would expect that models trained from scratch using language-specific tokenizers would achieve better results. We believe this result is related to the spoken nature of \texttt{MIAM} and further investigations are left as future work. Recent work \cite{rust2020good} has demonstrated that pretrained language models with language-specific tokenizers achieve better results than those using multilingual tokenizers. This result could explain the higher accuracy achieved by the language-specific versions of $MUG$ compared to $mMUG$.
\\\noindent We additionally observe that some language-specific versions of BERT achieve lower results (\textit{e.g} \texttt{Dihana}, \texttt{Loria}) than the multilingual version which could suggest that these pretrained BERT might be less carefully trained than the multilingual one; in the next part of the analysis we will only use multilingual tokenizers. 
\\\noindent\textbf{Overall, pretrained models achieve better results.} Contrarily to what can be observed in some syntactic tagging tasks \cite{zhang2018language}, for \texttt{DA} tagging pretrained models achieve consistently better results on the full benchmark. This result of multilingual models confirms what is observed with monolingual data (see \citet{mehri2019pretraining}): pretraining is an efficient method to build accurate dialog sequence labellers.
\\\noindent\textbf{Comparison of pretraining losses} In \autoref{tab:multilingual_da} we dissect the relative improvement brought by the different parts of the code-switched inspired losses and the architecture to better understand the relative importance of each component. Similarly to \citet{chapuis2020hierarchical}, we see that the hierarchical pretraining on spoken data (see $mMUG$) improves over the $mBERT$ model. Interestingly, we observe that the monolingual pretraining works slightly better compared to the multilingual pretraining when training using the same loss. This result surprising results might be attributed to the limited size of our models \cite{karthikeyan2019cross}.  
\\ We see that in both cases, introducing a loss with aligned multilingual conversations ($MMUG$ or $TMUG$) induces a performance gain (+$1.5\%$). This suggests that our pretraining with the new losses better captures the data distribution. By comparing the results of $mMUG+TMUG$ with $mMUG$, we observe that the addition of cross-lingual generation during pretraining helps. A marginal gain is induced when using $MMUG$ over $TMUG$, thus we believe that the improvement of   $mMUG+MMUG$ over $mMUG$ can mainly be attributed to the cross-lingual generation part. Interestingly, we observe that the combination of all losses out-performs the other models which suggests that different losses model different patterns present in the data. 
\subsubsection{Inconsistency Identification}\label{ssec:da_labelling}
In this section, we follow \citet{mehri2019pretraining} and evaluate our pretrained representations on \texttt{II} with a monolingual context. A random guess identifies the inconsistency by randomly selecting an index in $[1,T]$ which corresponds to an accuracy of $20\%$ (as we have set $T=5$). \autoref{tab:mono_inconsistency} gathers the results. Similarly conclusion than in \autoref{ssec:da_labelling} can be drawn: pretrained models achieve better results and the best performing model is obtained with $mMUG$+$MMUG$+$TMUG$.

\subsubsection{Next utterance retrieval}\label{ssec:da_labelling}
In this section, we evaluate our representations on \texttt{NUR} using a monolingual input context. As we use $9$ distractors, a random classifier would achieve $0.10$ for R@1, $0.20$ for R@2 and $0.50$ for R@5. The results are presented in \autoref{tab:mono_nru}. When comparing the accuracy obtained by the baselines models (\textit{e.g} mBERT, mBERT (4-layers) and $\mathcal{HR}$) and our model using the contextual losses at the context level for pretraining (\textit{i.e} $MUG$, $TMUG$ and $MMUG$) we observe a consistent improvement.

\textbf{Takeaways} Across all the three considered tasks, we observe that the models pretrained with our losses achieve better performances. We believe it is indicative of the validity of our pretraining.

\subsection{Multilingual input context}
In this section, we present the results on the downstream tasks with multilingual input context.
\subsubsection{\hspace{-0.3em}Multilingual inconsistency identification}
\autoref{tab:multi_inconsistency} gathers the results for the \texttt{mII} with bilingual input context. As previously a random baseline would achieve an accuracy of $20\%$. As expected predicting inconsistency with bilingual context is more challenging than with a monolingual context: we observe a drop in performance of around 15\% for all methods including the multilingual BERT. Our results confirm the observation of \citet{winata2021multilingual}: multilingual pretraining does not guarantee good performance in code switched data. However, we observe that the losses, by exposing the model with bilingual context, obtain a large boost (absolute improvement of 6\% which correspond to a relative boost of more than 20\%). We also observe that $MUG$+$MMUG$+$TMUG$ outperforms $mBERT$ on all pairs, with fewer parameters.

\subsubsection{Multilingual next utterance retrieval} 
The results on bilingual context for \texttt{mNUR} are presented in \autoref{tab:multi_nru}. \texttt{mNUR} is more challenging than \texttt{NUR}. Overall, we observe a strong gain in performance when exposing the model to bilingual context (gain over 9\% absolute point in R@5).
\textbf{Takeaways}: These results show that our code-switched inspired losses help to learn better representations in a particularly effective way in the case of multilingual input context. 

\begin{table}[!ht]
    \centering
    \resizebox{0.45\textwidth}{!}{\begin{tabular}{c|ccccc|c}\hline
                 & de & en & es & fr & it & Avg \\\hline
                  $mBERT$                 & \underline{44.6} & \underline{42.9} & \underline{43.7} & \underline{43.5}  & \underline{42.3} & \underline{43.4}       \\
  $mBERT$   (4-layers)     & \underline{44.6} & 42.1 & \underline{43.7} & 42.5  & 41.4 & 42.9        \\
                 $m\mathcal{HR}$ & 44.1 & 42.0 & 40.4 & 41.3  & 41.2 & 41.8      \\ 
    \hline
    \small{$mMUG$} & 45.2 & 43.5 & 45.1 & 43.1  & 42.7 & 43.9    \\
    \small{$mMUG+TMUG$}          & 48.2 & 42.6 & \textbf{47.7} & 44.6  & 44.3 & 45.5 \\
\small{$mMUG+MMUG$}  & \textbf{49.6} & \textbf{43.8} & 46.1 & \textbf{46.2}  & 43.3 & 45.8         \\
\small{$mMUG+TMUG+MMUG$}   & 49.1 & 43.4 & 46.2 & 45.9  & \textbf{45.1} & \textbf{46.0}          \\
\hline
\end{tabular}}
   \caption{Results on the \texttt{II} task with monolingual input context. On this task the accuracy is reported.}
    \label{tab:mono_inconsistency}
    \vspace{-0.5cm}
\end{table}
\begin{table*}[]
    \centering
    \resizebox{0.8\textwidth}{!}{\begin{tabular}{cc|ccccc|c}\hline
     & Toke.               & \texttt{VM2}  & \texttt{Map Task} & \texttt{Dihana} &  \texttt{Loria} & \texttt{Ilisten}  & Total \\\hline
    BERT & lang              & \underline{54.7} & \underline{66.4}  & 86.0    & 50.2& 74.9    & 66.4\\
    BERT - $4layers$  & lang & 52.8 & 66.2  & 85.8    & 55.2& \underline{76.2}    & 67.2\\
    $\mathcal{HR}$ + CRF & lang               & 49.7 & 63.1  & 85.8    & 73.4& 75.2    & 69.4\\
    $\mathcal{HR}$ + MLP & lang              & 51.3 & 63.0  & 85.6    & 58.9& 75.0    & 66.8\\
    $MUG$ \cite{chapuis2020hierarchical} & lang         & 54.0 & \underline{66.4} & \underline{99.0}   & \underline{79.0} & 74.8  & \underline{74.6} \\
    \hline\hline
    mBERT & multi             & \underline{53.2} & \underline{66.4}  & \underline{98.7}    & \underline{76.2}& 74.9   & \underline{73.8} \\
    mBERT - $4layers$ & multi  & 52.7 & 66.2  & 98.0    & 75.1& 75.0   & 73.4\\
    $m\mathcal{HR}$ + CRF & multi               & 49.8 & 65.2  & 97.6    & 75.2& \underline{76.0}   & 72.8 \\
    $m\mathcal{HR}$ + MLP & multi              & 51.0 & 65.7  & 97.8    & 75.2& \underline{76.0}    & 73.1\\\hline
    $mMUG$  & multi         & 53.0 & 67.3 & 98.3  & 78.5 & 74.0  & 74.2\\
    $mMUG+TMUG$  & multi      & 54.8 & \textbf{67.4} & 99.1   & \textbf{80.8} & 74.9  & 75.4\\
    $mMUG+MMUG$  & multi      & \textbf{56.2} & \textbf{67.4} & 99.0   & 78.9 & \textbf{77.6}  & 75.8 \\
      $mMUG+TMUG+MMUG$  & multi & \textbf{56.2} & 66.7 & \textbf{99.3}   & 80.7 & 77.0  & \textbf{76.0}\\
    \hline\end{tabular}}
    \caption{Accuracy of pretrained and baseline encoders on \texttt{MIAM}. Models are divided in three groups: hierarchical transformer encoders pretrained using our custom losses, baselines (see \autoref{ssec:baseline_details}) using either multilingual or language specific tokenizer.
    \textit{Toke.} stands for the type of tokenizer: \textit{multi} and \textit{lang} denotes a pretrained tokenizer on multilingual and language specific data respectively. When using $lang$ tokenizer, $MUG$ pretraining and finetuning are performed on the same language.} 
    \label{tab:multilingual_da}

\end{table*}

\begin{table*}[]
    \centering
    \resizebox{\textwidth}{!}{    \begin{tabular}{c|cccccccccc|c}\hline
                 & de-en & de-es & de-fr & de-it & en-es & en-fr & en-it & es-fr & es-it & fr-it & Avg \\\hline
                 mBERT                         & 31.2 & 28.0 & 28.0 & 27.6 & 28.4 & 33.0 & 32.1 & 35.1 & 31.0 & 28.7 & 30.3 \\
 mBERT  (4-layers)             & 30.7 & 28.7 & 28.2 & 27.1 & 28.7 & 33.1 & 30.9 & 35.1 & 30.1 & 28.1 & 30.1 \\
                  $m\mathcal{HR}$        & 28.7 & 27.9 & 26.9 & 27.3 & 25.5 & 25.1 & 30.6 & 34.3 & 30.0 & 26.8 & 28.3 \\ 
                  \hline

    $mMUG$                     & 34.5 & 30.1 & 30.1 & 27.7 & 28.2 & 33.1 & 32.1 & 35.4 & 32.0 & 29.5 & 31.2 \\
                      $mMUG+TMUG$           & 34.0 & 32.0 & 32.2 & 29.1 & 28.3 & 32.9 & 32.4 & 35.1 & 33.0 & 29.3 & 31.8  \\
                          $mMUG+MMUG$ & 35.1 & 33.8 & \textbf{34.0} & 30.1 & 29.4 & 32.8 & 32.6 & 36.1 & 33.9 & 31.6 &  32.9\\
        
         $mMUG+TMUG+MMUG$            & \textbf{35.7} & \textbf{34.0} & 32.5 & \textbf{31.4} & \textbf{30.1} & \textbf{33.6} & \textbf{33.9} & \textbf{36.2} & \textbf{34.0} & \textbf{32.1} & \textbf{33.4} \\
    \hline\end{tabular}}
   \caption{Results on the \texttt{mII} task with bilingual input context.}
    \label{tab:multi_inconsistency} 
\end{table*}

\begin{table*}[]
    \centering
    \resizebox{\textwidth}{!}{ \begin{tabular}{c|ccc|ccc|ccc|ccc|ccc}\hline
                 & \multicolumn{3}{c}{de} & \multicolumn{3}{c}{en} & \multicolumn{3}{c}{es} & \multicolumn{3}{c}{fr} & \multicolumn{3}{c}{it}  \\\hline
                                &R@5 &R@2 & R@1&R@5 &R@2 & R@1&R@5 &R@2 & R@1&R@5 &R@2 & R@1&R@5 &R@2 & R@1 \\\hline
                                
 mBERT                                                          & 65.1& 27.1&20.1     &62.1&26.1&16.8    &62.4&24.8&15.3     &63.9&22.9&13.4     &66.1&27.8&16.9   \\
 mBERT  (4-layers)                                              & 65.1& 27.5&20.2     &61.4&25.6&15.1    &62.3&24.6&15.9     &63.4&22.8&12.9     &65.6&27.4&15.8   \\
         $m\mathcal{HR}$                                                  & 65.0& 27.1&20.0     &60.3&25.0&15.2    &61.0&23.9&14.7     &63.0&22.9&13.0     &65.4&27.3&15.8   \\
         \hline
       $mMUG$  & 66.9& 28.0&20.0     &65.9&26.4&16.3    &66.7&26.4&16.4     &66.2&25.2&17.2     &68.9&28.9&17.2   \\    
         
                        $mMUG+TMUG$  & 67.2& \textbf{28.2}&20.1     &68.3&\textbf{29.8}&17.5    &69.0&26.9&17.3     &67.1&\textbf{25.4}&17.3     &69.9&29.4&18.6   \\
                       $mMUG+MMUG$  & 66.9& 28.1&20.7     &68.1&26.7&18.0    &68.7&26.9&17.5     &67.2&25.2&\textbf{17.4}     &69.7&29.4&18.6   \\
                 $mMUG+TMUG+MMUG$  & \textbf{68.3}& 27.4&\textbf{21.2}     &\textbf{68.9}&27.8&\textbf{18.3}    &\textbf{69.3}&\textbf{27.1}&\textbf{17.9}     &\textbf{67.4}&25.3&\textbf{17.4}     &\textbf{70.2}&\textbf{30.0}&\textbf{18.7}   \\
    \hline\end{tabular}}
    \caption{Results on the \texttt{NUR} task with monolingual input context. R@N stands for recall at $N$.} 
    \label{tab:mono_nru}
\end{table*}

\begin{table*}[]
    \centering
    \resizebox{\textwidth}{!}{ \begin{tabular}{c|ccc|ccc|ccc|ccc|ccc}\hline
                 & \multicolumn{3}{c}{de-en} & \multicolumn{3}{c}{de-es} & \multicolumn{3}{c}{de-fr} & \multicolumn{3}{c}{de-it} & \multicolumn{3}{c}{en-es}  \\\hline
                &R@5 &R@2 & R@1&R@5 &R@2 & R@1&R@5 &R@2 & R@1&R@5 &R@2 & R@1&R@5 &R@2 & R@1 \\\hline
                mBERT                        &54.4 &27.0 &11.6    &55.9 &24.8 &11.9    &57.9 &24.2 &12.9    &57.5&23.9  &13.0    &55.4&25.6 &13.0  \\
 mBERT  (4-layers)            &54.1 &26.5 &11.9    &55.7 &24.8 &\textbf{12.4}    &57.2 &24.1 &12.4    &57.0&23.5  &13.1    &55.6&23.1 &12.9  \\
         $m\mathcal{HR}$                &52.1 &25.5 &12.1    &54.9 &14.6 &10.7    &56.1 &22.9 &11.3    &56.9&24.9  &13.0    &53.9&23.7 &12.8\\
           
\hline
         $mMUG$                    &59.7 &25.2 &11.5    &61.2 &26.2 &11.6    &60.7 &25.3 &13.8    &61.6&\textbf{26.4}  &11.9    &62.1&23.9 &13.10   \\
         $mMUG+TMUG$               &59.8 &26.2 &12.1    &62.7 &29.0 &10.7    &61.9 &27.3 &13.9    &63.2&26.3  &12.6    &63.1&28.4 &14.0   \\
       $mMUG+MMUG$                 &59.8 &27.2 &12.1    &62.7 &28.1 &11.6    &60.7 &24.8 &\textbf{14.4}    &62.7&26.1  &\textbf{13.8}    &63.4&28.2 &14.7  \\
      
      $mMUG+TMUG+MMUG$       &\textbf{61.0} &\textbf{28.2} &\textbf{13.1}    &\textbf{63.2} &\textbf{29.1} &11.7    &\textbf{62.1} &\textbf{28.7} &14.1    &\textbf{63.4}&26.3  &12.9    &\textbf{64.3}&\textbf{29.4} &\textbf{15.2}  \\

\hline\hline
         
     &    \multicolumn{3}{c}{en-fr} & \multicolumn{3}{c}{en-it} & \multicolumn{3}{c}{es-fr} & \multicolumn{3}{c}{es-it} & \multicolumn{3}{c}{fr-it} \\\hline
                &R@5 &R@2 & R@1&R@5 &R@2 & R@1&R@5 &R@2 & R@1&R@5 &R@2 & R@1&R@5 &R@2 & R@1\\\hline
                
 mBERT                       &57.9&25.4 &12.3 &    57.1&23.5 &12.1 &    57.8&27.9 &12.2 &    54.2&22.1 &11.2 &    58.1&22.9 &12.5    \\
 mBERT  (4-layers)           &57.8&23.2 &12.1 &    57.1&23.4 &11.9 &    57.1&27.6 &12.1 &    55.1&22.0 &11.1 &    58.9&22.6 &12.7    \\
         $m\mathcal{HR}$               &55.9&20.9 &11.6 &    56.8&22.9 &11.8 &    54.9&27.0 &12.0 &    53.9&21.0 &11.6 &    56.1&21.9 &11.4    \\
    \hline
            $mMUG$          &61.9&24.9 &12.9 &    61.4&27.6 &11.9 &    64.6&29.7 &13.9 &    59.0&24.2 &13.4 &    59.7&23.6 &12.2    \\
               $mMUG+TMUG$              &62.9&25.2 &14.3 &    62.7&27.8 &12.9 &    64.9&29.9 &13.8 &    \textbf{60.1}&25.1 &13.5 &    61.5&25.8 &13.1    \\
    $mMUG+MMUG$            &63.9&26.3 &\textbf{14.7} &    61.5&27.6 &13.1 &    65.0&30.2 &13.1 &    \textbf{60.1}&25.3 &12.9 &    \textbf{63.1}&\textbf{25.9} &13.6    \\
            $mMUG+TMUG+MMUG$   &\textbf{64.0}&\textbf{26.7} &14.1 &    \textbf{63.5}&\textbf{28.7} &\textbf{13.7} &    \textbf{66.1}&\textbf{31.4} &\textbf{14.5} &    \textbf{60.1}&\textbf{25.5} &\textbf{13.6} &    \textbf{63.1}&\textbf{25.9} &\textbf{14.2}    \\
    \hline\end{tabular}}
    \caption{Results on the \texttt{mNUR} task with bilingual input context. } 
    \label{tab:multi_nru}
\end{table*}

\section{Conclusions}
In this work, we demonstrate that the new code-switched inspired losses help to learn representations for both monolingual and multilingual dialogs. This work is the first that explicitly includes code switching during pretraining to learn multilingual spoken dialog representations. In the future, we plan to further work on \texttt{OPS} to obtain fine-grained alignments (\textit{e.g} at the span and word levels) and enrich the definition of code-switching (currently limited at the utterance level). 
Lastly, when considering interactions with voice assistants and chatbots, users may not be able to express their intent in the language in which the voice assistant is programmed. Thus, we would like to strengthen our evaluation protocol by gathering a new \texttt{DA} benchmark with code-switched dialog to improve the multilingual evaluation. A possible future research direction includes focusing on emotion classification instead of dialog acts \cite{classif,jalalzai2020heavy}, extend our pretraining to multimodal data \cite{garcia-etal-2019-token,colombo2021improving} and use our model to obtain better results in sequence generation tasks (\textit{e.g} style transfer \cite{colombo2021novel,generic_emo}, automatic evaluation of natural language generation \cite{colombo2021automatic}). 

\section{Acknowledgments}
The research carried out in this paper has received funding from IBM, the French National Research Agency’s grant ANR-17-MAOI and the DSAIDIS chair at Telecom-Paris. This work was also granted access to the HPC resources of IDRIS under the allocation 2021-AP010611665 as well as under the project 2021-101838 made by GENCI. 
%



\bibliography{tacl2018}

\begin{thebibliography}{116}
\expandafter\ifx\csname natexlab\endcsname\relax\def\natexlab#1{#1}\fi

\bibitem[{Ahn et~al.(2020)Ahn, Jimenez, Tsvetkov, and Black}]{ahn2020code}
Emily Ahn, Cecilia Jimenez, Yulia Tsvetkov, and Alan~W Black. 2020.
\newblock What code-switching strategies are effective in dialog systems?
\newblock In \emph{Proceedings of the Society for Computation in Linguistics
  2020}, pages 213--222.

\bibitem[{Artetxe and Schwenk(2019)}]{artetxe2019massively}
Mikel Artetxe and Holger Schwenk. 2019.
\newblock Massively multilingual sentence embeddings for zero-shot
  cross-lingual transfer and beyond.
\newblock \emph{Transactions of the Association for Computational Linguistics},
  7:597--610.

\bibitem[{Auer(2013)}]{auer2013code}
Peter Auer. 2013.
\newblock \emph{Code-switching in conversation: Language, interaction and
  identity}.
\newblock Routledge.

\bibitem[{Banerjee et~al.(2018)Banerjee, Moghe, Arora, and
  Khapra}]{banerjee2018dataset}
Suman Banerjee, Nikita Moghe, Siddhartha Arora, and Mitesh~M Khapra. 2018.
\newblock A dataset for building code-mixed goal oriented conversation systems.
\newblock \emph{arXiv preprint arXiv:1806.05997}.

\bibitem[{Barahona et~al.(2012)Barahona, Lorenzo, and
  Gardent}]{barahona2012building}
Lina Maria~Rojas Barahona, Alejandra Lorenzo, and Claire Gardent. 2012.
\newblock Building and exploiting a corpus of dialog interactions between
  french speaking virtual and human agents.

\bibitem[{Basile and Novielli(2018)}]{basile2018overview}
Pierpaolo Basile and Nicole Novielli. 2018.
\newblock Overview of the evalita 2018 italian speech act labe ling (iliste n)
  task.
\newblock \emph{EVALITA Evaluation of NLP and Speech Tools for Italian}, 12:44.

\bibitem[{Bawa et~al.(2020)Bawa, Khadpe, Joshi, Bali, and
  Choudhury}]{bawa2020multilingual}
Anshul Bawa, Pranav Khadpe, Pratik Joshi, Kalika Bali, and Monojit Choudhury.
  2020.
\newblock Do multilingual users prefer chat-bots that code-mix? let's nudge and
  find out!
\newblock \emph{Proceedings of the ACM on Human-Computer Interaction},
  4(CSCW1):1--23.

\bibitem[{Bened{\i} et~al.(2006)Bened{\i}, Lleida, Varona, Castro, Galiano,
  Justo, L{\'o}pez, and Miguel}]{benedi2006design}
Jos{\'e}-Miguel Bened{\i}, Eduardo Lleida, Amparo Varona, Mar{\i}a-Jos{\'e}
  Castro, Isabel Galiano, Raquel Justo, I~L{\'o}pez, and Antonio Miguel. 2006.
\newblock Design and acquisition of a telephone spontaneous speech dialogue
  corpus in spanish: Dihana.
\newblock In \emph{Fifth International Conference on Language Resources and
  Evaluation (LREC)}, pages 1636--1639.

\bibitem[{Bothe et~al.(2018)Bothe, Weber, Magg, and Wermter}]{RNN_CTX3_Softmax}
Chandrakant Bothe, Cornelius Weber, Sven Magg, and Stefan Wermter. 2018.
\newblock A context-based approach for dialogue act recognition using simple
  recurrent neural networks.
\newblock \emph{CoRR}, abs/1805.06280.

\bibitem[{Cañete et~al.(2020)Cañete, Chaperon, Fuentes, Ho, Kang, and
  Pérez}]{CaneteCFP2020}
José Cañete, Gabriel Chaperon, Rodrigo Fuentes, Jou-Hui Ho, Hojin Kang, and
  Jorge Pérez. 2020.
\newblock Spanish pre-trained bert model and evaluation data.
\newblock In \emph{PML4DC at ICLR 2020}.

\bibitem[{Celikyilmaz et~al.(2020)Celikyilmaz, Clark, and
  Gao}]{celikyilmaz2020evaluation}
Asli Celikyilmaz, Elizabeth Clark, and Jianfeng Gao. 2020.
\newblock Evaluation of text generation: A survey.
\newblock \emph{arXiv preprint arXiv:2006.14799}.

\bibitem[{Chapuis et~al.(2020)Chapuis, Colombo, Manica, Labeau, and
  Clavel}]{chapuis2020hierarchical}
Emile Chapuis, Pierre Colombo, Matteo Manica, Matthieu Labeau, and Chlo{\'e}
  Clavel. 2020.
\newblock \href {https://doi.org/10.18653/v1/2020.findings-emnlp.239}
  {Hierarchical pre-training for sequence labelling in spoken dialog}.
\newblock In \emph{Findings of the Association for Computational Linguistics:
  EMNLP 2020}, pages 2636--2648, Online. Association for Computational
  Linguistics.

\bibitem[{Chen et~al.(2018{\natexlab{a}})Chen, Yang, Zhao, Cai, and
  He}]{crf_chen}
Zheqian Chen, Rongqin Yang, Zhou Zhao, Deng Cai, and Xiaofei He.
  2018{\natexlab{a}}.
\newblock Dialogue act recognition via crf-attentive structured network.
\newblock In \emph{The 41st International ACM SIGIR Conference on Research \&
  Development in Information Retrieval}, pages 225--234.

\bibitem[{Chen et~al.(2018{\natexlab{b}})Chen, Yang, Zhao, Cai, and
  He}]{LSTM_CRF}
Zheqian Chen, Rongqin Yang, Zhou Zhao, Deng Cai, and Xiaofei He.
  2018{\natexlab{b}}.
\newblock Dialogue act recognition via crf-attentive structured network.
\newblock In \emph{The 41st International ACM SIGIR Conference on Research \&
  Development in Information Retrieval}, pages 225--234. ACM.

\bibitem[{Chung et~al.(2014)Chung, Gulcehre, Cho, and Bengio}]{gru}
Junyoung Chung, Caglar Gulcehre, KyungHyun Cho, and Yoshua Bengio. 2014.
\newblock Empirical evaluation of gated recurrent neural networks on sequence
  modeling.
\newblock \emph{arXiv preprint arXiv:1412.3555}.

\bibitem[{Colombo et~al.(2021{\natexlab{a}})Colombo, Chapuis, Labeau, and
  Clavel}]{colombo2021improving}
Pierre Colombo, Emile Chapuis, Matthieu Labeau, and Chloe Clavel.
  2021{\natexlab{a}}.
\newblock Improving multimodal fusion via mutual dependency maximisation.
\newblock \emph{arXiv preprint arXiv:2109.00922}.

\bibitem[{Colombo et~al.(2020)Colombo, Chapuis, Manica, Vignon, Varni, and
  Clavel}]{colombo2020guiding}
Pierre Colombo, Emile Chapuis, Matteo Manica, Emmanuel Vignon, Giovanna Varni,
  and Chloe Clavel. 2020.
\newblock Guiding attention in sequence-to-sequence models for dialogue act
  prediction.
\newblock \emph{arXiv preprint arXiv:2002.08801}.

\bibitem[{Colombo et~al.(2021{\natexlab{b}})Colombo, Clavel, and
  Piantanida}]{colombo2021novel}
Pierre Colombo, Chloe Clavel, and Pablo Piantanida. 2021{\natexlab{b}}.
\newblock A novel estimator of mutual information for learning to disentangle
  textual representations.
\newblock \emph{arXiv preprint arXiv:2105.02685}.

\bibitem[{Colombo et~al.(2021{\natexlab{c}})Colombo, Staerman, Clavel, and
  Piantanida}]{colombo2021automatic}
Pierre Colombo, Guillaume Staerman, Chloe Clavel, and Pablo Piantanida.
  2021{\natexlab{c}}.
\newblock Automatic text evaluation through the lens of wasserstein
  barycenters.
\newblock \emph{arXiv preprint arXiv:2108.12463}.

\bibitem[{Colombo et~al.(2019)Colombo, Witon, Modi, Kennedy, and
  Kapadia}]{generic_emo}
Pierre Colombo, Wojciech Witon, Ashutosh Modi, James Kennedy, and Mubbasir
  Kapadia. 2019.
\newblock Affect-driven dialog generation.
\newblock \emph{arXiv preprint arXiv:1904.02793}.

\bibitem[{Conneau et~al.(2019)Conneau, Khandelwal, Goyal, Chaudhary, Wenzek,
  Guzm{\'a}n, Grave, Ott, Zettlemoyer, and Stoyanov}]{conneau2019unsupervised}
Alexis Conneau, Kartikay Khandelwal, Naman Goyal, Vishrav Chaudhary, Guillaume
  Wenzek, Francisco Guzm{\'a}n, Edouard Grave, Myle Ott, Luke Zettlemoyer, and
  Veselin Stoyanov. 2019.
\newblock Unsupervised cross-lingual representation learning at scale.
\newblock \emph{arXiv preprint arXiv:1911.02116}.

\bibitem[{Coria and Pineda(2005)}]{coria2005predicting}
S~Coria and L~Pineda. 2005.
\newblock Predicting obligation dialogue acts from prosodic and speaker
  infomation.
\newblock \emph{Research on Computing Science (ISSN 1665-9899), Centro de
  Investigacion en Computacion, Instituto Politecnico Nacional, Mexico City}.

\bibitem[{Dai et~al.(2019)Dai, Yang, Yang, Carbonell, Le, and
  Salakhutdinov}]{dai2019transformer}
Zihang Dai, Zhilin Yang, Yiming Yang, Jaime Carbonell, Quoc~V Le, and Ruslan
  Salakhutdinov. 2019.
\newblock Transformer-xl: Attentive language models beyond a fixed-length
  context.
\newblock \emph{arXiv preprint arXiv:1901.02860}.

\bibitem[{Devlin et~al.(2018)Devlin, Chang, Lee, and Toutanova}]{bert}
Jacob Devlin, Ming-Wei Chang, Kenton Lee, and Kristina Toutanova. 2018.
\newblock Bert: Pre-training of deep bidirectional transformers for language
  understanding.
\newblock \emph{arXiv preprint arXiv:1810.04805}.

\bibitem[{Dinkar et~al.(2020)Dinkar, Colombo, Labeau, and
  Clavel}]{dinkar2020importance}
Tanvi Dinkar, Pierre Colombo, Matthieu Labeau, and Chlo{\'e} Clavel. 2020.
\newblock The importance of fillers for text representations of speech
  transcripts.
\newblock \emph{arXiv preprint arXiv:2009.11340}.

\bibitem[{Dinkar et~al.(2018)Dinkar, Vasilescu, Pelachaud, and
  Clavel}]{dinkar2018disfluencies}
Tanvi Dinkar, Ioana Vasilescu, Catherine Pelachaud, and Chlo{\'e} Clavel. 2018.
\newblock Disfluencies and teaching strategies in social interactions between a
  pedagogical agent and a student: Background and challenges.
\newblock In \emph{SEMDIAL 2018 (AixDial), The 22nd workshop on the Semantics
  and Pragmatics of Dialogue}, pages 188--191. Laurent Pr{\'e}vot, Magalie Ochs
  and Beno{\^\i}t Favre.

\bibitem[{Duplessis et~al.(2017)Duplessis, Charras, Letard, Ligozat, and
  Rosset}]{duplessis2017utterance}
Guillaume~Dubuisson Duplessis, Franck Charras, Vincent Letard, Anne-Laure
  Ligozat, and Sophie Rosset. 2017.
\newblock Utterance retrieval based on recurrent surface text patterns.
\newblock In \emph{European Conference on Information Retrieval}, pages
  199--211. Springer.

\bibitem[{Dziri et~al.(2019)Dziri, Kamalloo, Mathewson, and
  Zaiane}]{dziri2019evaluating}
Nouha Dziri, Ehsan Kamalloo, Kory~W Mathewson, and Osmar Zaiane. 2019.
\newblock Evaluating coherence in dialogue systems using entailment.
\newblock \emph{arXiv preprint arXiv:1904.03371}.

\bibitem[{Eriguchi et~al.(2018)Eriguchi, Johnson, Firat, Kazawa, and
  Macherey}]{eriguchi2018zero}
Akiko Eriguchi, Melvin Johnson, Orhan Firat, Hideto Kazawa, and Wolfgang
  Macherey. 2018.
\newblock Zero-shot cross-lingual classification using multilingual neural
  machine translation.
\newblock \emph{arXiv preprint arXiv:1809.04686}.

\bibitem[{Escolano et~al.(2020)Escolano, Costa-juss{\`a}, Fonollosa, and
  Artetxe}]{escolano2020training}
Carlos Escolano, Marta~R Costa-juss{\`a}, Jos{\'e}~AR Fonollosa, and Mikel
  Artetxe. 2020.
\newblock Training multilingual machine translation by alternately freezing
  language-specific encoders-decoders.
\newblock \emph{arXiv preprint arXiv:2006.01594}.

\bibitem[{Fairchild and Van~Hell(2017)}]{fairchild2017determiner}
Sarah Fairchild and Janet~G Van~Hell. 2017.
\newblock Determiner-noun code-switching in spanish heritage speakers.
\newblock \emph{Bilingualism: Language and Cognition}, 20(1):150--161.

\bibitem[{Faruqui and Dyer(2014)}]{faruqui2014improving}
Manaal Faruqui and Chris Dyer. 2014.
\newblock Improving vector space word representations using multilingual
  correlation.
\newblock In \emph{Proceedings of the 14th Conference of the European Chapter
  of the Association for Computational Linguistics}, pages 462--471.

\bibitem[{Feng et~al.(2020)Feng, Yang, Cer, Arivazhagan, and
  Wang}]{feng2020language}
Fangxiaoyu Feng, Yinfei Yang, Daniel Cer, Naveen Arivazhagan, and Wei Wang.
  2020.
\newblock Language-agnostic bert sentence embedding.
\newblock \emph{arXiv preprint arXiv:2007.01852}.

\bibitem[{Finch and Choi(2020)}]{finch2020towards}
Sarah~E Finch and Jinho~D Choi. 2020.
\newblock Towards unified dialogue system evaluation: A comprehensive analysis
  of current evaluation protocols.
\newblock \emph{arXiv preprint arXiv:2006.06110}.

\bibitem[{Fraser and Gilbert(1991)}]{fraser1991simulating}
Norman~M Fraser and G~Nigel Gilbert. 1991.
\newblock Simulating speech systems.
\newblock \emph{Computer Speech \& Language}, 5(1):81--99.

\bibitem[{Garcia et~al.(2019)Garcia, Colombo, d{'}Alch{\'e} Buc, Essid, and
  Clavel}]{garcia-etal-2019-token}
Alexandre Garcia, Pierre Colombo, Florence d{'}Alch{\'e} Buc, Slim Essid, and
  Chlo{\'e} Clavel. 2019.
\newblock \href {https://doi.org/10.18653/v1/D19-1556} {From the {T}oken to the
  {R}eview: {A} {H}ierarchical {M}ultimodal {A}pproach to {O}pinion {M}ining}.
\newblock In \emph{Proceedings of the 2019 Conference on Empirical Methods in
  Natural Language Processing and the 9th International Joint Conference on
  Natural Language Processing (EMNLP-IJCNLP)}, pages 5539--5548, Hong Kong,
  China. Association for Computational Linguistics.

\bibitem[{Ghosal et~al.(2019)Ghosal, Majumder, Poria, Chhaya, and
  Gelbukh}]{weighted_preproc}
Deepanway Ghosal, Navonil Majumder, Soujanya Poria, Niyati Chhaya, and
  Alexander Gelbukh. 2019.
\newblock Dialoguegcn: A graph convolutional neural network for emotion
  recognition in conversation.
\newblock \emph{arXiv preprint arXiv:1908.11540}.

\bibitem[{Godfrey et~al.(1992)Godfrey, Holliman, and McDaniel}]{datase_swda}
John~J. Godfrey, Edward~C. Holliman, and Jane McDaniel. 1992.
\newblock Switchboard: Telephone speech corpus for research and development.
\newblock In \emph{Proceedings of the 1992 IEEE International Conference on
  Acoustics, Speech and Signal Processing - Volume 1}, ICASSP’92, page
  517–520, USA. IEEE Computer Society.

\bibitem[{Gouws et~al.(2015)Gouws, Bengio, and Corrado}]{gouws2015bilbowa}
Stephan Gouws, Yoshua Bengio, and Greg Corrado. 2015.
\newblock Bilbowa: Fast bilingual distributed representations without word
  alignments.

\bibitem[{Grosjean and Li(2013)}]{grosjean2013psycholinguistics}
Fran{\c{c}}ois Grosjean and Ping Li. 2013.
\newblock \emph{The psycholinguistics of bilingualism}.
\newblock John Wiley \& Sons.

\bibitem[{Gumperz(1982)}]{gumperz1982discourse}
John~J Gumperz. 1982.
\newblock \emph{Discourse strategies}.
\newblock 1. Cambridge University Press.

\bibitem[{Henderson et~al.(2019)Henderson, Casanueva, Mrk{\v{s}}i{\'c}, Su,
  Wen, and Vuli{\'c}}]{henderson2019convert}
Matthew Henderson, I{\~n}igo Casanueva, Nikola Mrk{\v{s}}i{\'c}, Pei-Hao Su,
  Tsung-Hsien Wen, and Ivan Vuli{\'c}. 2019.
\newblock Convert: Efficient and accurate conversational representations from
  transformers.
\newblock \emph{arXiv preprint arXiv:1911.03688}.

\bibitem[{Hendrycks and Gimpel(2016)}]{gelu}
Dan Hendrycks and Kevin Gimpel. 2016.
\newblock Gaussian error linear units (gelus).
\newblock \emph{arXiv preprint arXiv:1606.08415}.

\bibitem[{Heredia and Altarriba(2001)}]{heredia2001bilingual}
Roberto~R Heredia and Jeanette Altarriba. 2001.
\newblock Bilingual language mixing: Why do bilinguals code-switch?
\newblock \emph{Current Directions in Psychological Science}, 10(5):164--168.

\bibitem[{Hochreiter and Schmidhuber(1997)}]{lstm}
Sepp Hochreiter and J{\"u}rgen Schmidhuber. 1997.
\newblock Long short-term memory.
\newblock \emph{Neural computation}, 9(8):1735--1780.

\bibitem[{Ipsic et~al.(1999)Ipsic, Pavesic, Mihelic, and
  Noth}]{ipsic1999multilingual}
Ivo Ipsic, Nikola Pavesic, France Mihelic, and Elmar Noth. 1999.
\newblock Multilingual spoken dialog system.
\newblock In \emph{ISIE'99. Proceedings of the IEEE International Symposium on
  Industrial Electronics (Cat. No. 99TH8465)}, volume~1, pages 183--187. IEEE.

\bibitem[{Jalalzai et~al.(2020)Jalalzai, Colombo, Clavel, Gaussier, Varni,
  Vignon, and Sabourin}]{jalalzai2020heavy}
Hamid Jalalzai, Pierre Colombo, Chlo{\'e} Clavel, Eric Gaussier, Giovanna
  Varni, Emmanuel Vignon, and Anne Sabourin. 2020.
\newblock Heavy-tailed representations, text polarity classification \& data
  augmentation.
\newblock \emph{arXiv preprint arXiv:2003.11593}.

\bibitem[{Jiao et~al.(2019)Jiao, Yin, Shang, Jiang, Chen, Li, Wang, and
  Liu}]{tiny}
Xiaoqi Jiao, Yichun Yin, Lifeng Shang, Xin Jiang, Xiao Chen, Linlin Li, Fang
  Wang, and Qun Liu. 2019.
\newblock Tinybert: Distilling bert for natural language understanding.
\newblock \emph{arXiv preprint arXiv:1909.10351}.

\bibitem[{Joshi et~al.(2020)Joshi, Santy, Budhiraja, Bali, and
  Choudhury}]{joshi2020state}
Pratik Joshi, Sebastin Santy, Amar Budhiraja, Kalika Bali, and Monojit
  Choudhury. 2020.
\newblock The state and fate of linguistic diversity and inclusion in the nlp
  world.
\newblock \emph{arXiv preprint arXiv:2004.09095}.

\bibitem[{Karthikeyan et~al.(2019)Karthikeyan, Wang, Mayhew, and
  Roth}]{karthikeyan2019cross}
Kaliyaperumal Karthikeyan, Zihan Wang, Stephen Mayhew, and Dan Roth. 2019.
\newblock Cross-lingual ability of multilingual bert: An empirical study.
\newblock In \emph{International Conference on Learning Representations}.

\bibitem[{Kay et~al.(1992)Kay, Norvig, and Gawron}]{kay1992verbmobil}
Martin Kay, Peter Norvig, and Mark Gawron. 1992.
\newblock \emph{Verbmobil: A translation system for face-to-face dialog}.
\newblock University of Chicago Press.

\bibitem[{Keizer et~al.(2002)Keizer, op~den Akker, and
  Nijholt}]{bayesian_dialog}
Simon Keizer, Rieks op~den Akker, and Anton Nijholt. 2002.
\newblock Dialogue act recognition with bayesian networks for dutch dialogues.
\newblock In \emph{Proceedings of the Third SIGdial Workshop on Discourse and
  Dialogue}.

\bibitem[{Khanpour et~al.(2016)Khanpour, Guntakandla, and
  Nielsen}]{LSTM_softmax}
Hamed Khanpour, Nishitha Guntakandla, and Rodney Nielsen. 2016.
\newblock Dialogue act classification in domain-independent conversations using
  a deep recurrent neural network.
\newblock In \emph{COLING}.

\bibitem[{Khanuja et~al.(2020)Khanuja, Dandapat, Srinivasan, Sitaram, and
  Choudhury}]{khanuja2020gluecos}
Simran Khanuja, Sandipan Dandapat, Anirudh Srinivasan, Sunayana Sitaram, and
  Monojit Choudhury. 2020.
\newblock Gluecos: An evaluation benchmark for code-switched nlp.
\newblock \emph{arXiv preprint arXiv:2004.12376}.

\bibitem[{Kingma and Ba(2015)}]{adam}
Diederik~P. Kingma and Jimmy Ba. 2015.
\newblock \href {https://arxiv.org/pdf/1412.6980.pdf} {{A}dam: {A} {M}ethod for
  {S}tochastic {O}ptimization}.
\newblock \emph{CoRR}, abs/1412.6980.

\bibitem[{Koiso et~al.(1998)Koiso, Horiuchi, Tutiya, Ichikawa, and
  Den}]{koiso1998analysis}
Hanae Koiso, Yasuo Horiuchi, Syun Tutiya, Akira Ichikawa, and Yasuharu Den.
  1998.
\newblock An analysis of turn-taking and backchannels based on prosodic and
  syntactic features in japanese map task dialogs.
\newblock \emph{Language and speech}, 41(3-4):295--321.

\bibitem[{Kudo and Richardson(2018)}]{kudo2018sentencepiece}
Taku Kudo and John Richardson. 2018.
\newblock \href {https://doi.org/10.18653/v1/D18-2012} {{S}entence{P}iece: {A}
  {S}imple and {L}anguage {I}ndependent {S}ubword {T}okenizer and {D}etokenizer
  for {N}eural {T}ext {P}rocessing}.
\newblock In \emph{Proceedings of the 2018 Conference on Empirical Methods in
  Natural Language Processing: System Demonstrations (EMNLP)}, pages 66--71,
  Brussels, Belgium. Association for Computational Linguistics.

\bibitem[{Lafferty et~al.(2001)Lafferty, McCallum, and Pereira}]{crf}
John Lafferty, Andrew McCallum, and Fernando~CN Pereira. 2001.
\newblock Conditional random fields: Probabilistic models for segmenting and
  labeling sequence data.

\bibitem[{Lample and Conneau(2019)}]{lample2019cross}
Guillaume Lample and Alexis Conneau. 2019.
\newblock Cross-lingual language model pretraining.
\newblock \emph{arXiv preprint arXiv:1901.07291}.

\bibitem[{Lan et~al.(2019)Lan, Chen, Goodman, Gimpel, Sharma, and
  Soricut}]{albert}
Zhenzhong Lan, Mingda Chen, Sebastian Goodman, Kevin Gimpel, Piyush Sharma, and
  Radu Soricut. 2019.
\newblock Albert: A lite bert for self-supervised learning of language
  representations.
\newblock \emph{arXiv preprint arXiv:1909.11942}.

\bibitem[{Le et~al.(2019)Le, Vial, Frej, Segonne, Coavoux, Lecouteux, Allauzen,
  Crabb{\'e}, Besacier, and Schwab}]{flaubert}
Hang Le, Lo{\"\i}c Vial, Jibril Frej, Vincent Segonne, Maximin Coavoux,
  Benjamin Lecouteux, Alexandre Allauzen, Beno{\^\i}t Crabb{\'e}, Laurent
  Besacier, and Didier Schwab. 2019.
\newblock Flaubert: Unsupervised language model pre-training for french.
\newblock \emph{arXiv preprint arXiv:1912.05372}.

\bibitem[{Li et~al.(2018)Li, Lin, Collinson, Li, and Chen}]{sota_swda_1}
Ruizhe Li, Chenghua Lin, Matthew Collinson, Xiao Li, and Guanyi Chen. 2018.
\newblock \href {http://arxiv.org/abs/1810.09154} {A dual-attention
  hierarchical recurrent neural network for dialogue act classification}.
\newblock \emph{CoRR}, abs/1810.09154.

\bibitem[{Li et~al.(2017)Li, Su, Shen, Li, Cao, and Niu}]{dataset_dailydialog}
Yanran Li, Hui Su, Xiaoyu Shen, Wenjie Li, Ziqiang Cao, and Shuzi Niu. 2017.
\newblock \href {http://arxiv.org/abs/1710.03957} {Dailydialog: A manually
  labelled multi-turn dialogue dataset}.

\bibitem[{Lin et~al.(2017)Lin, Feng, Santos, Yu, Xiang, Zhou, and
  Bengio}]{self_attention}
Zhouhan Lin, Minwei Feng, Cicero Nogueira~dos Santos, Mo~Yu, Bing Xiang, Bowen
  Zhou, and Yoshua Bengio. 2017.
\newblock A structured self-attentive sentence embedding.
\newblock \emph{arXiv preprint arXiv:1703.03130}.

\bibitem[{Lison and Tiedemann(2016)}]{opensub}
Pierre Lison and J{\"o}rg Tiedemann. 2016.
\newblock Opensubtitles2016: Extracting large parallel corpora from movie and
  tv subtitles.

\bibitem[{Liu et~al.(2020)Liu, Gu, Goyal, Li, Edunov, Ghazvininejad, Lewis, and
  Zettlemoyer}]{liu2020multilingual}
Yinhan Liu, Jiatao Gu, Naman Goyal, Xian Li, Sergey Edunov, Marjan
  Ghazvininejad, Mike Lewis, and Luke Zettlemoyer. 2020.
\newblock Multilingual denoising pre-training for neural machine translation.
\newblock \emph{Transactions of the Association for Computational Linguistics},
  8:726--742.

\bibitem[{Liu et~al.(2019)Liu, Ott, Goyal, Du, Joshi, Chen, Levy, Lewis,
  Zettlemoyer, and Stoyanov}]{roberta}
Yinhan Liu, Myle Ott, Naman Goyal, Jingfei Du, Mandar Joshi, Danqi Chen, Omer
  Levy, Mike Lewis, Luke Zettlemoyer, and Veselin Stoyanov. 2019.
\newblock Roberta: A robustly optimized bert pretraining approach.
\newblock \emph{arXiv preprint arXiv:1907.11692}.

\bibitem[{Loshchilov and Hutter(2017)}]{adamW}
Ilya Loshchilov and Frank Hutter. 2017.
\newblock Decoupled weight decay regularization.
\newblock \emph{arXiv preprint arXiv:1711.05101}.

\bibitem[{Lowe et~al.(2015)Lowe, Pow, Serban, and Pineau}]{ubuntu}
Ryan Lowe, Nissan Pow, Iulian Serban, and Joelle Pineau. 2015.
\newblock \href {http://arxiv.org/abs/1506.08909} {The ubuntu dialogue corpus:
  {A} large dataset for research in unstructured multi-turn dialogue systems}.
\newblock \emph{CoRR}, abs/1506.08909.

\bibitem[{Lowe et~al.(2016)Lowe, Serban, Noseworthy, Charlin, and
  Pineau}]{lowe2016evaluation}
Ryan Lowe, Iulian~V Serban, Mike Noseworthy, Laurent Charlin, and Joelle
  Pineau. 2016.
\newblock On the evaluation of dialogue systems with next utterance
  classification.
\newblock \emph{arXiv preprint arXiv:1605.05414}.

\bibitem[{Mehri et~al.(2019)Mehri, Razumovsakaia, Zhao, and
  Eskenazi}]{mehri2019pretraining}
Shikib Mehri, Evgeniia Razumovsakaia, Tiancheng Zhao, and Maxine Eskenazi.
  2019.
\newblock Pretraining methods for dialog context representation learning.
\newblock \emph{arXiv preprint arXiv:1906.00414}.

\bibitem[{Mikolov et~al.(2013)Mikolov, Le, and
  Sutskever}]{mikolov2013exploiting}
Tomas Mikolov, Quoc~V Le, and Ilya Sutskever. 2013.
\newblock Exploiting similarities among languages for machine translation.
\newblock \emph{arXiv preprint arXiv:1309.4168}.

\bibitem[{Milroy et~al.(1995)}]{milroy1995one}
James Milroy et~al. 1995.
\newblock \emph{One speaker, two languages: Cross-disciplinary perspectives on
  code-switching}.
\newblock Cambridge University Press.

\bibitem[{Mitkov(2014)}]{corner2}
Ruslan Mitkov. 2014.
\newblock \emph{Anaphora resolution}.
\newblock Routledge.

\bibitem[{Olguin and Cort{\'e}s(2006)}]{olguin2006predicting}
Sergio Rafael~Coria Olguin and Luis Albreto~Pineda Cort{\'e}s. 2006.
\newblock Predicting dialogue acts from prosodic information.
\newblock In \emph{International Conference on Intelligent Text Processing and
  Computational Linguistics}, pages 355--365. Springer.

\bibitem[{Parekh et~al.(2020)Parekh, Ahn, Tsvetkov, and
  Black}]{parekh2020understanding}
Tanmay Parekh, Emily Ahn, Yulia Tsvetkov, and Alan~W Black. 2020.
\newblock Understanding linguistic accommodation in code-switched human-machine
  dialogues.
\newblock In \emph{Proceedings of the 24th Conference on Computational Natural
  Language Learning}, pages 565--577.

\bibitem[{Paszke et~al.(2017)Paszke, Gross, Chintala, Chanan, Yang, DeVito,
  Lin, Desmaison, Antiga, and Lerer}]{pytorch}
Adam Paszke, Sam Gross, Soumith Chintala, Gregory Chanan, Edward Yang, Zachary
  DeVito, Zeming Lin, Alban Desmaison, Luca Antiga, and Adam Lerer. 2017.
\newblock Automatic differentiation in pytorch.

\bibitem[{Poplack(1980)}]{poplack1980sometimes}
Shana Poplack. 1980.
\newblock Sometimes i’ll start a sentence in spanish y termino en espanol:
  toward a typology of code-switching1.

\bibitem[{Poria et~al.(2018)Poria, Hazarika, Majumder, Naik, Cambria, and
  Mihalcea}]{weighted_no}
Soujanya Poria, Devamanyu Hazarika, Navonil Majumder, Gautam Naik, Erik
  Cambria, and Rada Mihalcea. 2018.
\newblock Meld: A multimodal multi-party dataset for emotion recognition in
  conversations.
\newblock \emph{arXiv preprint arXiv:1810.02508}.

\bibitem[{Post et~al.(2013)Post, Kumar, Lopez, Karakos, Callison-Burch, and
  Khudanpur}]{post2013improved}
Matt Post, Gaurav Kumar, Adam Lopez, Damianos Karakos, Chris Callison-Burch,
  and Sanjeev Khudanpur. 2013.
\newblock Improved speech-to-text translation with the {F}isher and {C}allhome
  {S}panish--{E}nglish speech translation corpus.
\newblock In \emph{Proceedings of the International Workshop on Spoken Language
  Translation (IWSLT)}, Heidelberg, Germany.

\bibitem[{Pratapa et~al.(2018)Pratapa, Choudhury, and
  Sitaram}]{pratapa2018word}
Adithya Pratapa, Monojit Choudhury, and Sunayana Sitaram. 2018.
\newblock Word embeddings for code-mixed language processing.
\newblock In \emph{Proceedings of the 2018 conference on empirical methods in
  natural language processing}, pages 3067--3072.

\bibitem[{Qi et~al.(2021)Qi, Gong, Yan, Xu, Yao, Zhou, Cheng, Jiang, Chen,
  Zhang et~al.}]{qi2021prophetnet}
Weizhen Qi, Yeyun Gong, Yu~Yan, Can Xu, Bolun Yao, Bartuer Zhou, Biao Cheng,
  Daxin Jiang, Jiusheng Chen, Ruofei Zhang, et~al. 2021.
\newblock Prophetnet-x: Large-scale pre-training models for english, chinese,
  multi-lingual, dialog, and code generation.
\newblock \emph{arXiv preprint arXiv:2104.08006}.

\bibitem[{Ribeiro et~al.(2019{\natexlab{a}})Ribeiro, Ribeiro, and
  de~Matos}]{dihana_1}
Eug{\'e}nio Ribeiro, Ricardo Ribeiro, and David~Martins de~Matos.
  2019{\natexlab{a}}.
\newblock Hierarchical multi-label dialog act recognition on spanish data.
\newblock \emph{arXiv preprint arXiv:1907.12316}.

\bibitem[{Ribeiro et~al.(2019{\natexlab{b}})Ribeiro, Ribeiro, and
  de~Matos}]{ribeiro2019multilingual}
Eug{\'e}nio Ribeiro, Ricardo Ribeiro, and David~Martins de~Matos.
  2019{\natexlab{b}}.
\newblock A multilingual and multidomain study on dialog act recognition using
  character-level tokenization.
\newblock \emph{Information}, 10(3):94.

\bibitem[{Ruder et~al.(2019)Ruder, Vuli{\'c}, and S{\o}gaard}]{ruder2019survey}
Sebastian Ruder, Ivan Vuli{\'c}, and Anders S{\o}gaard. 2019.
\newblock A survey of cross-lingual word embedding models.
\newblock \emph{Journal of Artificial Intelligence Research}, 65:569--631.

\bibitem[{Rust et~al.(2020)Rust, Pfeiffer, Vuli{\'c}, Ruder, and
  Gurevych}]{rust2020good}
Phillip Rust, Jonas Pfeiffer, Ivan Vuli{\'c}, Sebastian Ruder, and Iryna
  Gurevych. 2020.
\newblock How good is your tokenizer? on the monolingual performance of
  multilingual language models.
\newblock \emph{arXiv preprint arXiv:2012.15613}.

\bibitem[{Sankar et~al.(2019{\natexlab{a}})Sankar, Subramanian, Pal, Chandar,
  and Bengio}]{sankar2019neural}
Chinnadhurai Sankar, Sandeep Subramanian, Christopher Pal, Sarath Chandar, and
  Yoshua Bengio. 2019{\natexlab{a}}.
\newblock Do neural dialog systems use the conversation history effectively? an
  empirical study.
\newblock \emph{arXiv preprint arXiv:1906.01603}.

\bibitem[{Sankar et~al.(2019{\natexlab{b}})Sankar, Subramanian, Pal, Chandar,
  and Bengio}]{incosistency_dialog}
Chinnadhurai Sankar, Sandeep Subramanian, Christopher Pal, Sarath Chandar, and
  Yoshua Bengio. 2019{\natexlab{b}}.
\newblock Do neural dialog systems use the conversation history effectively? an
  empirical study.
\newblock \emph{arXiv preprint arXiv:1906.01603}.

\bibitem[{Sankoff and Poplack(1981)}]{sankoff1981formal}
David Sankoff and Shana Poplack. 1981.
\newblock A formal grammar for code-switching.
\newblock \emph{Research on Language \& Social Interaction}, 14(1):3--45.

\bibitem[{Saraclar and Sproat(2004)}]{saraclar2004lattice}
Murat Saraclar and Richard Sproat. 2004.
\newblock Lattice-based search for spoken utterance retrieval.
\newblock In \emph{Proceedings of the Human Language Technology Conference of
  the North American Chapter of the Association for Computational Linguistics:
  HLT-NAACL 2004}, pages 129--136.

\bibitem[{Schatzmann et~al.(2005)Schatzmann, Georgila, and
  Young}]{schatzmann2005quantitative}
Jost Schatzmann, Kallirroi Georgila, and Steve Young. 2005.
\newblock Quantitative evaluation of user simulation techniques for spoken
  dialogue systems.
\newblock In \emph{6th SIGdial Workshop on DISCOURSE and DIALOGUE}.

\bibitem[{Schweter(2020)}]{stefan_schweter_2020_4263142}
Stefan Schweter. 2020.
\newblock \href {https://doi.org/10.5281/zenodo.4263142} {Italian bert and
  electra models}.

\bibitem[{Sener and Koltun(2018)}]{multi_loss_opt}
Ozan Sener and Vladlen Koltun. 2018.
\newblock Multi-task learning as multi-objective optimization.
\newblock In \emph{Advances in Neural Information Processing Systems}, pages
  527--538.

\bibitem[{Shriberg et~al.(2004)Shriberg, Dhillon, Bhagat, Ang, and
  Carvey}]{dataset_mrda}
Elizabeth Shriberg, Raj Dhillon, Sonali Bhagat, Jeremy Ang, and Hannah Carvey.
  2004.
\newblock \href {https://www.aclweb.org/anthology/W04-2319} {The {ICSI} meeting
  recorder dialog act ({MRDA}) corpus}.
\newblock In \emph{Proceedings of the 5th {SIG}dial Workshop on Discourse and
  Dialogue at {HLT}-{NAACL} 2004}, pages 97--100, Cambridge, Massachusetts,
  USA. Association for Computational Linguistics.

\bibitem[{Srivastava et~al.(2014)Srivastava, Hinton, Krizhevsky, Sutskever, and
  Salakhutdinov}]{dropout}
Nitish Srivastava, Geoffrey Hinton, Alex Krizhevsky, Ilya Sutskever, and Ruslan
  Salakhutdinov. 2014.
\newblock Dropout: a simple way to prevent neural networks from overfitting.
\newblock \emph{The journal of machine learning research}, 15(1):1929--1958.

\bibitem[{Stolcke et~al.(2000{\natexlab{a}})Stolcke, Ries, Coccaro, Shriberg,
  Bates, Jurafsky, Taylor, Martin, Ess-Dykema, and Meteer}]{hmm_dialog}
Andreas Stolcke, Klaus Ries, Noah Coccaro, Elizabeth Shriberg, Rebecca Bates,
  Daniel Jurafsky, Paul Taylor, Rachel Martin, Carol~Van Ess-Dykema, and Marie
  Meteer. 2000{\natexlab{a}}.
\newblock Dialogue act modeling for automatic tagging and recognition of
  conversational speech.
\newblock \emph{Computational linguistics}, 26(3):339--373.

\bibitem[{Stolcke et~al.(2000{\natexlab{b}})Stolcke, Ries, Coccaro, Shriberg,
  Bates, Jurafsky, Taylor, Martin, Ess-Dykema, and Meteer}]{handcrafted_HMM}
Andreas Stolcke, Klaus Ries, Noah Coccaro, Elizabeth Shriberg, Rebecca Bates,
  Daniel Jurafsky, Paul Taylor, Rachel Martin, Carol~Van Ess-Dykema, and Marie
  Meteer. 2000{\natexlab{b}}.
\newblock Dialogue act modeling for automatic tagging and recognition of
  conversational speech.
\newblock \emph{Computational linguistics}, 26(3):339--373.

\bibitem[{Stymne et~al.(2020)}]{stymne2020evaluating}
Sara Stymne et~al. 2020.
\newblock Evaluating word embeddings for indonesian--english code-mixed text
  based on synthetic data.
\newblock In \emph{Proceedings of the The 4th Workshop on Computational
  Approaches to Code Switching}, pages 26--35.

\bibitem[{Surendran and Levow(2006)}]{svm_dialog}
Dinoj Surendran and Gina-Anne Levow. 2006.
\newblock Dialog act tagging with support vector machines and hidden markov
  models.
\newblock In \emph{Ninth International Conference on Spoken Language
  Processing}.

\bibitem[{Tan and Joty(2021)}]{tan2021code}
Samson Tan and Shafiq Joty. 2021.
\newblock Code-mixing on sesame street: Dawn of the adversarial polyglots.
\newblock \emph{arXiv preprint arXiv:2103.09593}.

\bibitem[{Taylor(1953)}]{cloze}
Wilson~L Taylor. 1953.
\newblock “cloze procedure”: A new tool for measuring readability.
\newblock \emph{Journalism quarterly}, 30(4):415--433.

\bibitem[{Thornbury and Slade(2006)}]{context}
Scott Thornbury and Diana Slade. 2006.
\newblock \emph{Conversation: From description to pedagogy}.
\newblock Cambridge University Press.

\bibitem[{Vaswani et~al.(2017)Vaswani, Shazeer, Parmar, Uszkoreit, Jones,
  Gomez, Kaiser, and Polosukhin}]{attention_is}
Ashish Vaswani, Noam Shazeer, Niki Parmar, Jakob Uszkoreit, Llion Jones,
  Aidan~N Gomez, {\L}ukasz Kaiser, and Illia Polosukhin. 2017.
\newblock Attention is all you need.
\newblock In \emph{Advances in neural information processing systems}, pages
  5998--6008.

\bibitem[{Williams et~al.(2014)Williams, Henderson, Raux, Thomson, Black, and
  Ramachandran}]{corner1}
Jason~D Williams, Matthew Henderson, Antoine Raux, Blaise Thomson, Alan Black,
  and Deepak Ramachandran. 2014.
\newblock The dialog state tracking challenge series.
\newblock \emph{AI Magazine}, 35(4):121--124.

\bibitem[{Winata et~al.(2021)Winata, Cahyawijaya, Liu, Lin, Madotto, and
  Fung}]{winata2021multilingual}
Genta~Indra Winata, Samuel Cahyawijaya, Zihan Liu, Zhaojiang Lin, Andrea
  Madotto, and Pascale Fung. 2021.
\newblock Are multilingual models effective in code-switching?
\newblock \emph{arXiv preprint arXiv:2103.13309}.

\bibitem[{Witon et~al.(2018)Witon, Colombo, Modi, and Kapadia}]{classif}
Wojciech Witon, Pierre Colombo, Ashutosh Modi, and Mubbasir Kapadia. 2018.
\newblock Disney at iest 2018: Predicting emotions using an ensemble.
\newblock In \emph{Proceedings of the 9th Workshop on Computational Approaches
  to Subjectivity, Sentiment and Social Media Analysis}, pages 248--253.

\bibitem[{Wolf et~al.(2019)Wolf, Debut, Sanh, Chaumond, Delangue, Moi, Cistac,
  Rault, Louf, Funtowicz, and Brew}]{Wolf2019HuggingFacesTS}
Thomas Wolf, Lysandre Debut, Victor Sanh, Julien Chaumond, Clement Delangue,
  Anthony Moi, Pierric Cistac, Tim Rault, R'emi Louf, Morgan Funtowicz, and
  Jamie Brew. 2019.
\newblock Huggingface's transformers: State-of-the-art natural language
  processing.
\newblock \emph{ArXiv}, abs/1910.03771.

\bibitem[{Wolf et~al.(2020)Wolf, Lhoest, von Platen, Jernite, Drame, Plu,
  Chaumond, Delangue, Ma, Thakur, Patil, Davison, Scao, Sanh, Xu, Patry,
  McMillan-Major, Brandeis, Gugger, Lagunas, Debut, Funtowicz, Moi, Rush,
  Schmidd, Cistac, Muštar, Boudier, and Tordjmann}]{2020HuggingFace-datasets}
Thomas Wolf, Quentin Lhoest, Patrick von Platen, Yacine Jernite, Mariama Drame,
  Julien Plu, Julien Chaumond, Clement Delangue, Clara Ma, Abhishek Thakur,
  Suraj Patil, Joe Davison, Teven~Le Scao, Victor Sanh, Canwen Xu, Nicolas
  Patry, Angie McMillan-Major, Simon Brandeis, Sylvain Gugger, François
  Lagunas, Lysandre Debut, Morgan Funtowicz, Anthony Moi, Sasha Rush, Philipp
  Schmidd, Pierric Cistac, Victor Muštar, Jeff Boudier, and Anna Tordjmann.
  2020.
\newblock Datasets.
\newblock \emph{GitHub. Note: https://github.com/huggingface/datasets}, 1.

\bibitem[{Wu et~al.(2016)Wu, Schuster, Chen, Le, Norouzi, Macherey, Krikun,
  Cao, Gao, Macherey et~al.}]{wordpiece}
Yonghui Wu, Mike Schuster, Zhifeng Chen, Quoc~V Le, Mohammad Norouzi, Wolfgang
  Macherey, Maxim Krikun, Yuan Cao, Qin Gao, Klaus Macherey, et~al. 2016.
\newblock Google's neural machine translation system: Bridging the gap between
  human and machine translation.
\newblock \emph{arXiv preprint arXiv:1609.08144}.

\bibitem[{Xue et~al.(2020)Xue, Constant, Roberts, Kale, Al-Rfou, Siddhant,
  Barua, and Raffel}]{xue2020mt5}
Linting Xue, Noah Constant, Adam Roberts, Mihir Kale, Rami Al-Rfou, Aditya
  Siddhant, Aditya Barua, and Colin Raffel. 2020.
\newblock \href {http://arxiv.org/abs/2010.11934} {{mT5}: A massively
  multilingual pre-trained text-to-text transformer}.

\bibitem[{Yang et~al.(2019)Yang, Dai, Yang, Carbonell, Salakhutdinov, and
  Le}]{xlnet}
Zhilin Yang, Zihang Dai, Yiming Yang, Jaime Carbonell, Russ~R Salakhutdinov,
  and Quoc~V Le. 2019.
\newblock Xlnet: Generalized autoregressive pretraining for language
  understanding.
\newblock In \emph{Advances in neural information processing systems}, pages
  5754--5764.

\bibitem[{Zhang and Bowman(2018)}]{zhang2018language}
Kelly Zhang and Samuel Bowman. 2018.
\newblock Language modeling teaches you more than translation does: Lessons
  learned through auxiliary syntactic task analysis.
\newblock In \emph{Proceedings of the 2018 EMNLP Workshop BlackboxNLP:
  Analyzing and Interpreting Neural Networks for NLP}, pages 359--361.

\bibitem[{Zhang et~al.(2019{\natexlab{a}})Zhang, Kishore, Wu, Weinberger, and
  Artzi}]{zhang2019bertscore}
Tianyi Zhang, Varsha Kishore, Felix Wu, Kilian~Q Weinberger, and Yoav Artzi.
  2019{\natexlab{a}}.
\newblock Bertscore: Evaluating text generation with bert.
\newblock \emph{arXiv preprint arXiv:1904.09675}.

\bibitem[{Zhang et~al.(2019{\natexlab{b}})Zhang, Wei, and
  Zhou}]{zhang2019hibert}
Xingxing Zhang, Furu Wei, and Ming Zhou. 2019{\natexlab{b}}.
\newblock Hibert: Document level pre-training of hierarchical bidirectional
  transformers for document summarization.
\newblock \emph{arXiv preprint arXiv:1905.06566}.

\bibitem[{Zhang et~al.(2019{\natexlab{c}})Zhang, Li, Song, Zhang, and
  Wang}]{accuracy_fscore}
Yazhou Zhang, Qiuchi Li, Dawei Song, Peng Zhang, and Panpan Wang.
  2019{\natexlab{c}}.
\newblock Quantum-inspired interactive networks for conversational sentiment
  analysis.

\bibitem[{Zhang et~al.(2019{\natexlab{d}})Zhang, Sun, Galley, Chen, Brockett,
  Gao, Gao, Liu, and Dolan}]{zhang2019dialogpt}
Yizhe Zhang, Siqi Sun, Michel Galley, Yen-Chun Chen, Chris Brockett, Xiang Gao,
  Jianfeng Gao, Jingjing Liu, and Bill Dolan. 2019{\natexlab{d}}.
\newblock Dialogpt: Large-scale generative pre-training for conversational
  response generation.
\newblock \emph{arXiv preprint arXiv:1911.00536}.

\end{thebibliography}
\bibliographystyle{acl_natbib}

\clearpage
\section{Additional details on evaluation corpus}

\subsection{\texttt{MIAM}: examples and diversity}
\begin{figure*}[h]
\centering
 \begin{subfigure}[b]{0.44\textwidth}
     \centering
     \includegraphics[width=\textwidth]{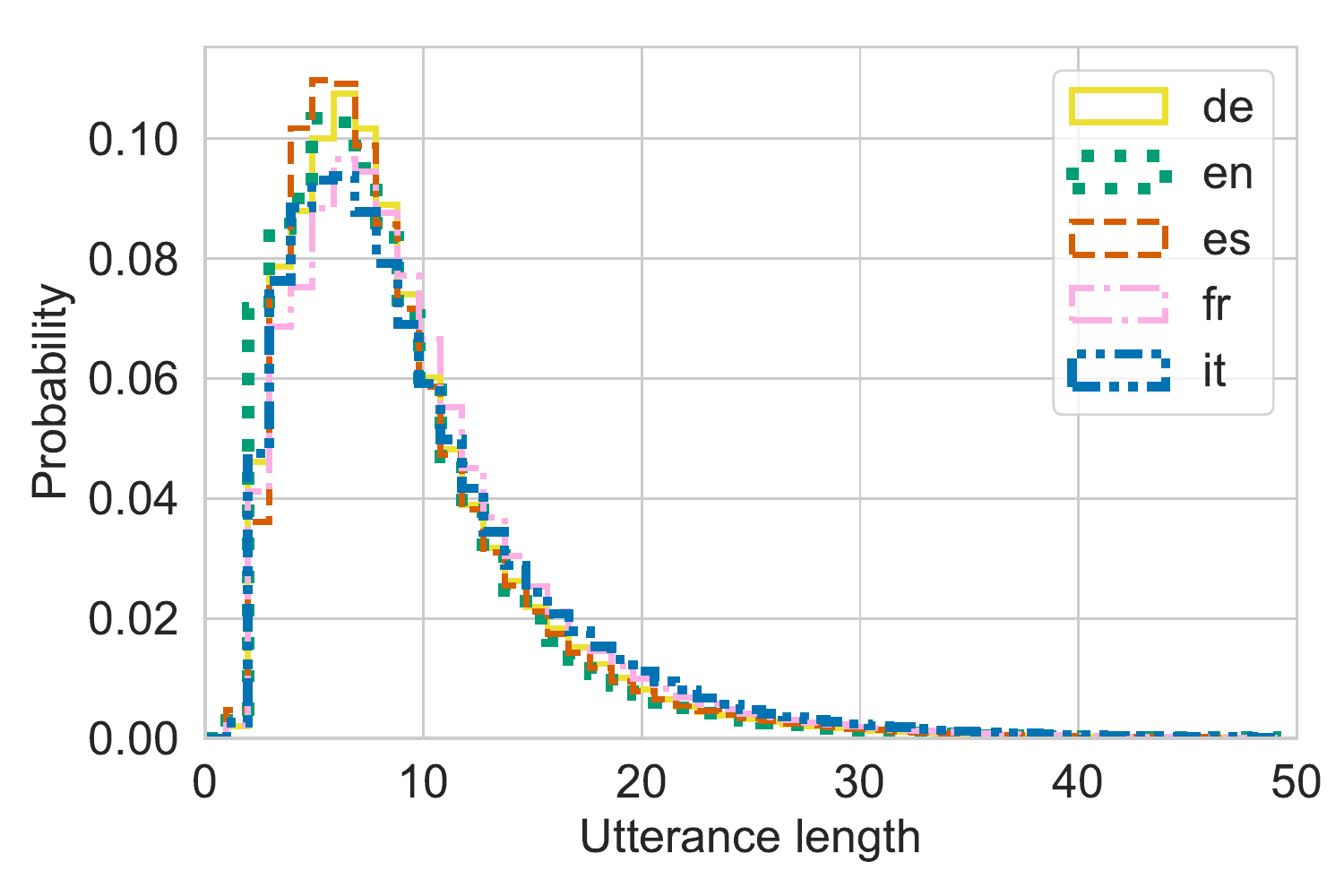}
 \end{subfigure}
 \begin{subfigure}[b]{0.44\textwidth}
     \centering
     \includegraphics[width=\textwidth]{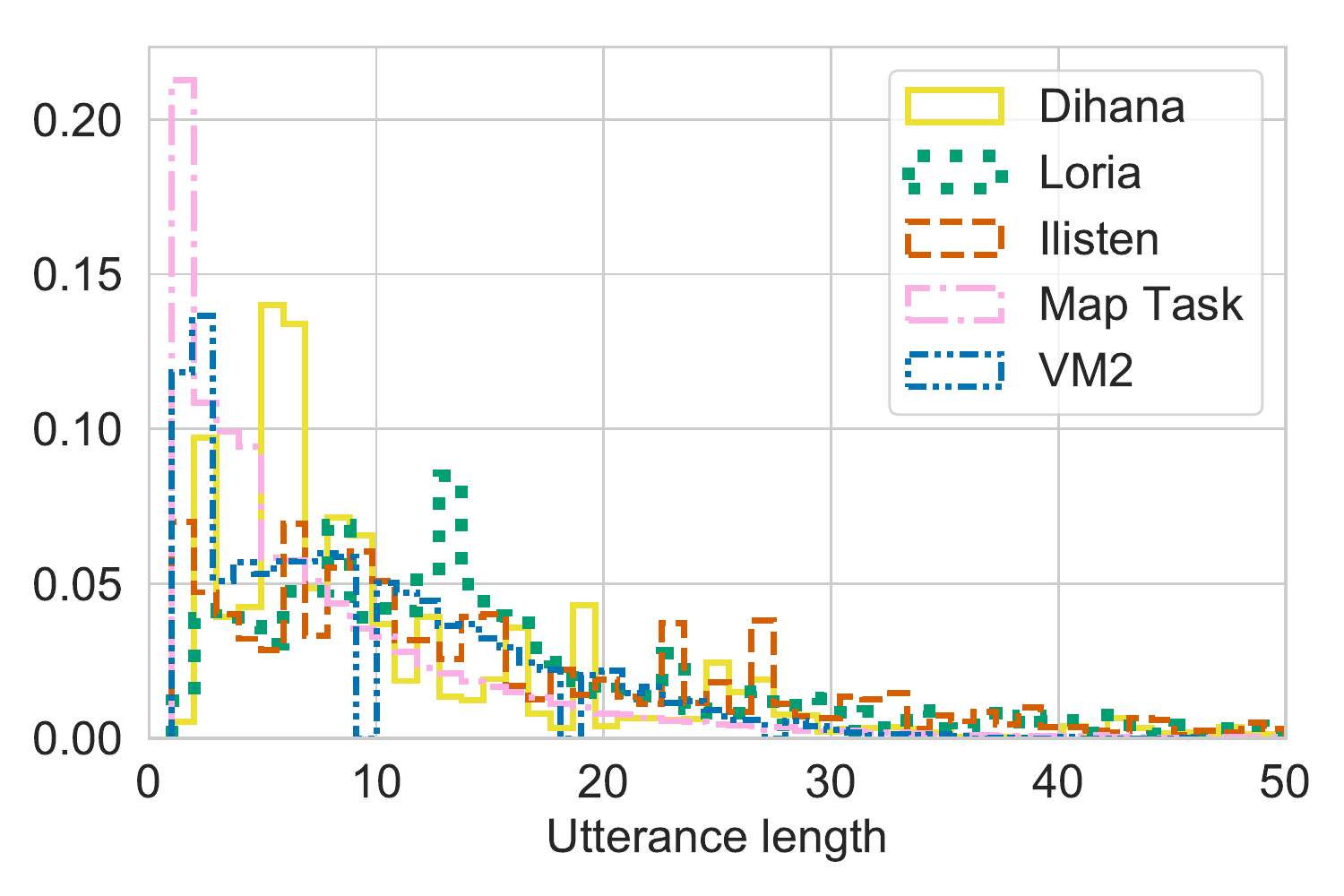}
 \end{subfigure}
 \caption{Histograms showing the utterance length for \texttt{OPS} (left) and \texttt{MIAM} (right).}
 \label{fig:utt_hist}
 
\end{figure*}

In this section we give more details on the \texttt{MIAM} benchmark. \autoref{tab:comp_ex} shows examples extracted from the benchmark. In \autoref{fig:utt_hist} we illustrate the diversity of the gathered corpora through the lens of utterance length. 
\begin{table}[!htb]
\centering
\resizebox{.47\textwidth}{!}{\begin{tabular}{c|c|c} 
\hline
Lang.&Utterances & \texttt{DA}  \\
\hline
\multirow{5}{*}{de}&\makecell{soll ich dann mit dem Hotel \\da dann die Buchung vereinbaren} &	\texttt{OFFER}\\
&ja das ist gut &	\texttt{FEED. POS.}\\
&das wäre toll &	\texttt{ACCEPT}\\
&dann kümmere ich mich um die Tickets &	\texttt{COMMIT}\\
&wunderbar &	\texttt{ACCEPT}\\\hline
\multirow{5}{*}{en}&how far underneath the diamond mine & \texttt{ASK}\\
&it's about an inch or so & \texttt{REPLY}\\
&right okay five inches right along & \texttt{ACK.}\\
&up along to near a r-- a ravine stuff thing & \texttt{ASK} \\
&no i don't have the ravine & \texttt{REPLY} \\\hline
\multirow{5}{*}{es}&¿ Qué día desea salir ?  &	\texttt{ASK}\\
&El diez de noviembre . &	\texttt{REQUEST}\\
&Quiere horarios de trenes a barcelona, &	\texttt{CONFIRM}\\
&¿ desde zaragoza ? &	\texttt{CONFIRM}\\
&Sí , por favor . &	\texttt{AFF.}\\\hline
\multirow{5}{*}{fr}&Bonjour &	\texttt{GREETINGS}\\
&Bonjour , je suis Sophia l'opérateur (...). &	\texttt{GREETINGS}\\
&Enchanté &	\texttt{GREETINGS}\\
&Qu'est ce que je peux faire pour vous ? &	\texttt{ASK}\\
&\makecell{J'ai besoins des informations sur \\ les composants de la manette.} &	\texttt{INFORMER}\\
\hline
\multirow{5}{*}{it}&mangio tre volte al giorno & \texttt{STATEMENT} \\
&Ti piace mangiare? & \texttt{QUESTION} \\
&abbastanza & \texttt{ANSWER} \\
&Che cosa hai mangiato per colazione? & \texttt{QUESTION} \\
&latte e biscotti & \texttt{STATEMENT} \\\hline
\end{tabular}}
\caption{Examples of dialogs labelled with \texttt{DA} taken from \texttt{MapTask}, \texttt{Dihana}, \texttt{VM2}, \texttt{Loria} and \texttt{Ilisten}. \texttt{AFF.} stands for affirmation, \texttt{FEED.} for feedback and \texttt{ACK.} for acknowledgement.}
\label{tab:comp_ex}
\end{table}

\subsection{Altering tasks difficulty} 
One of the interesting properties of \texttt{II}, \texttt{mII}, \texttt{NUR}, \texttt{mNUR} is the ability to alter the task difficulty in a controlled manner when sampling the negative utterances. For example, instead of randomly sampling the false utterances, the most similar to the true one as measured by a similarity metric \cite{zhang2019bertscore,celikyilmaz2020evaluation} could be chosen. This flexibility could allow increasing the difficulty of the task as models get better. 
\section{Experimental settings}

\subsection{Additional details on pretrained models}\label{ssec:additionnal_pretriained}
In this section, we gather additional details on the pretrained models (\textit{e.g} architectures, schema, hyperparameters).

\subsubsection{Pretraining losses}
\begin{figure*}[!h]
\centering
 \begin{subfigure}{0.45\textwidth}
     \centering
     \includegraphics[width=\textwidth]{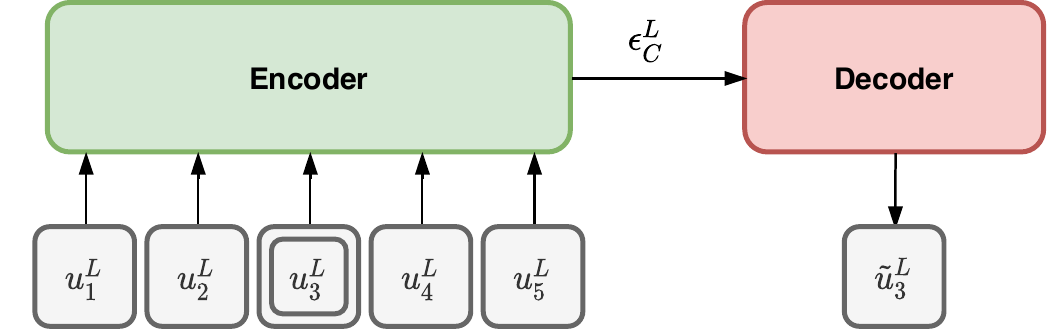}
     \caption{$MUG$}
     \label{fig:loss_1}
     \hfill
 \end{subfigure}
 \begin{subfigure}{0.45\textwidth}
     \centering
     \includegraphics[width=\textwidth]{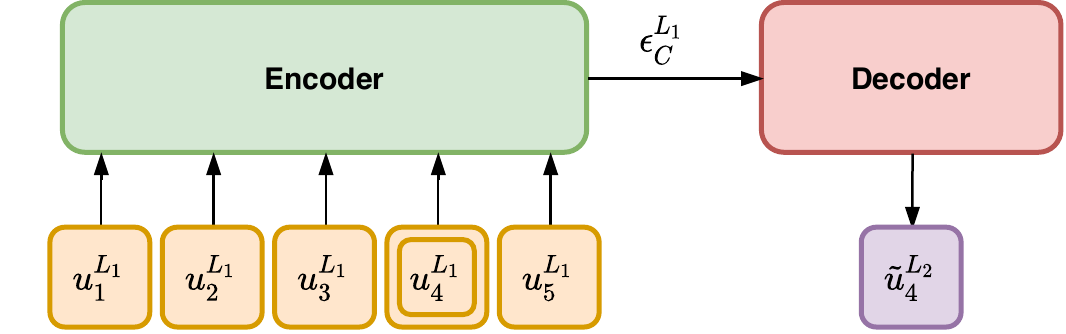}
     \caption{$TMUG$}
     \label{fig:loss_2}
     \hfill
 \end{subfigure}
 \begin{subfigure}{0.45\textwidth}
     \centering
     \includegraphics[width=\textwidth]{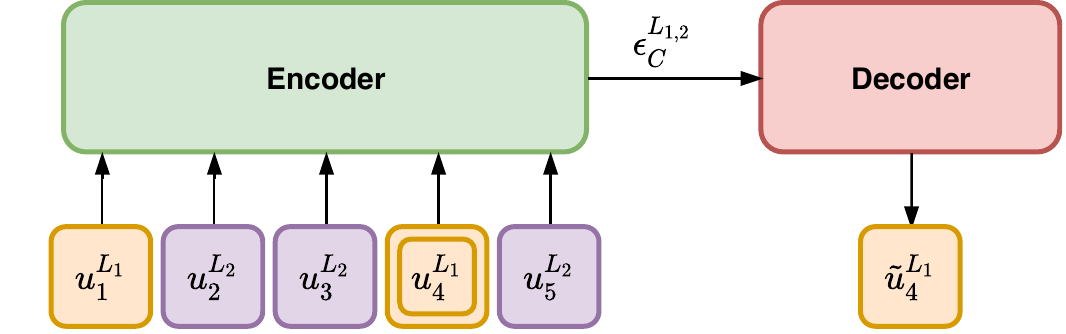}
     \caption{$MMUG$: prediction of $u_4^{L_1}$}
     \label{fig:loss_3}
 \end{subfigure}
 \begin{subfigure}{0.45\textwidth}
     \centering
     \includegraphics[width=\textwidth]{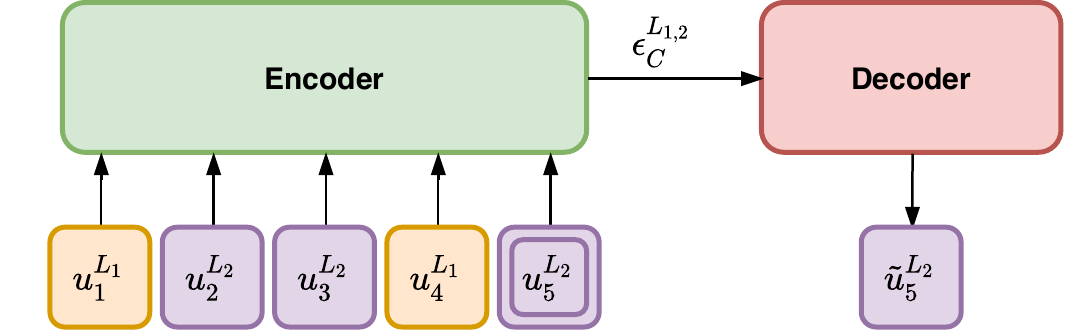}
     \caption{$MMUG$: prediction of $u_5^{L_3}$}
     \label{fig:loss_4}
 \end{subfigure}
 \caption{
 \ref{fig:loss_1} and \ref{fig:loss_2} illustrate pretraining losses using monolingual context. 
 \ref{fig:loss_2} and \ref{fig:loss_3} show two scenarios for the $MMUG$ loss using multilingual context. Double squares on the figure indicates the randomly selected utterance to predict.}
 \label{fig:losses}
\end{figure*}

\autoref{fig:losses} gives graphical examples for each monolingual and multilingual losses used.
\textbf{Choice of scaling factor in \autoref{eq:h_loss}}. In the case of multi-task setting, different losses may have different scales, making the optimization perform poorly. In that case, scaling factors or more advanced techniques \cite{multi_loss_opt} can be applied. As we did not observe such phenomena, all scaling factors are set to 1.

\subsubsection{Pretraining with generation} For both TMUG and MMUG, the model needs to be aware of the target language. Thus, the first token fed to the decoder indicates the target language (\textit{e.g} in English the corresponding id is  99, in Spanish 98). To avoid creating a discrepancy between pretraining objectives we also add this token for MUG.

\subsubsection{Choice of the multilingual encoder}
The two dominant approaches for multilingual systems involve either using a language-specific encoder \cite{escolano2020training} or one shared encoder across languages \cite{feng2020language,artetxe2019massively}. To reduce the number of learnt parameters, we rely on the second approach.


\subsubsection{Pretraining details}
Our model is pretrained on 4 NVIDIA V100 for 2 days (500k iterations) with a batch size of $256$. We use AdamW \cite{adam,adamW}  with $4000$ warmups steps \cite{attention_is}. During this stage, we do not perform any grid search. 

\subsection{Additional details on downstream task}
In this section, we gather additional details on downstream tasks (\textit{e.g} choice of pretrained encoders, choice of decoder and further details on the downstream tasks). 
\subsubsection{Pretrained encoders baseline}
\label{supp:pretrained_encoder}
The first group of pretrained encoders are based on BERT. A concatenation of utterances is fed to the model to obtain a conversation embedding.  For our language-specific models, we use the German BERT\footnote{https://deepset.ai/}, the original BERT for English, BETO \cite{CaneteCFP2020} for Spanish, Flaubert~\cite{flaubert} for French  and Italian BERT \citet{stefan_schweter_2020_4263142} for Italian. We rely on the multilingual BERT (mBERT)~\cite{bert}\footnote{https://github.com/google-research/bert/blob/master/multilingual.md} provided by the transformers library~\cite{Wolf2019HuggingFacesTS} implemented using the pytorch~\cite{pytorch} framework. For pretrained hierarchical transformers, we rely on the work of \citet{chapuis2020hierarchical} and for each considered language, we pretrain a language-specific encoder.

\subsubsection{Decoders}
Given the different nature of the proposed downstream tasks, we use various type of decoders.
\noindent\textbf{\texttt{DA} classification}: Methods to tackle sequence labelling on monolingual representations can be divided into two different classes. The first one perform classification on each utterance independently using Bayesian Networks \cite{bayesian_dialog}, SVMs \cite{svm_dialog} or HMMs \cite{handcrafted_HMM}. The second class, which achieves stronger results, leverages the adjacency utterances by using deep representations \cite{RNN_CTX3_Softmax,LSTM_softmax}. Sequence labelling can be improved when sufficiently many training points are available by modelling inter-tag dependencies using RNN-based decoders \cite{lstm,gru}, and CRFs \cite{crf,LSTM_CRF}. Thus, in this work, we choose to experiment with a MLP, a CRF and a RNN decoder based on GRU.  

\noindent\textbf{\texttt{II} and \texttt{mII} }: For this task, the context embedding $\mathcal{E}_{C_k}$ is fed to a MLP. Both the encoder and the MLP are trained to predict the inconsistent utterance index by minimising a cross-entropy loss. Formally, this task is formulated as a classification problem with $T$ classes. 

\noindent\textbf{\texttt{NUR} and \texttt{mNUR}}: For this task, we first compute the context embedding $\mathcal{E}_{C_k}$ then the candidate utterance $u^{L_c}_c$ is embedded using the either $f_\theta^u$ or a chosen encoder to obtain $\mathcal{E}_{u^{L_c}_c}$. Both representations are concatenated and given to a MLP. The architecture is trained to predict if the provided candidate utterance is a suitable next utterance by minimizing a binary cross-entropy. This experiment is similar to the one in \cite{ubuntu}.

\subsection{Additional details on models}\label{ssec:baseline_details}
In this section, we describe models used as well as details on the pretraining parameters.
In~\autoref{tab:archi_hyper} we report the main hyper-parameters used for our model pretraining. We used GELU~\cite{gelu} activations and the dropout rate \cite{dropout} is set to $0.1$. Although vanilla Transformers impose a fixed context size it can be relaxed \cite{dai2019transformer}. We follow~\citet{sankar2019neural,colombo2020guiding} and set $T=5$. We rely on the tokenizers provided by the HuggingFace library based on the SentencePiece \cite{kudo2018sentencepiece} and WordPiece \cite{wordpiece} algorithms. In all experiments, for our models relying on the $\mathcal{HT}$ we use the same architecture as the \texttt{SMALL} model from \citet{chapuis2020hierarchical} which contains ~80 millions parameters. Original BERT has 167 millions parameters and is pretrained using 16 TPUs during several days with over 500K iterations.

\begin{table}[ht]
    \centering
    \begin{tabular}{c|cc}\hline
     & Pretrained Encoder \\\hline
     Nbs of heads  & 6  \\
      $N_d$ &4 \\
      $N_u$ & 4\\
      $T$  & 50\\
      $C$  & 5\\
      $\mathcal{T}_d$ nbs of heads &6 \\
      Inner dimension &768 \\
  Model Dimension    &768 \\
$|\mathcal{V}|$  &105879 \\
  $\mathcal{T}_d$: Emb. size &768 \\
  $d_k$:&64 \\
  $d_v$: &64
    \end{tabular}
    \caption{Architecture hyperparameters used for the hierarchical pretraining.}
    \label{tab:archi_hyper}
\end{table}
\subsection{Training details} 
\label{sec:training_details}
For each task, the model is fine-tuned and dropout \cite{dropout} is set to $0.1$. The best learning rate is found in $\{0.01,0.001,0.0001\}$ and chosen based on the validation loss.

\section{Additional experiment: ablation study on pretraining data}
\label{sec:ablation_pretraining_data}
We showcase the difference between pretraining with spoken and written corpora. We compare $m\mathcal{H}\mathcal{T}(\theta_{written})$, a hierarchical encoder where each utterance is embedded using the representation of the \texttt{[CLS]} token given by the second layer of BERT, and $m\mathcal{H}\mathcal{T}_u(\theta_{spoken})$, a model pretrained on \texttt{OPS} using $\mathcal{L}^u$ only. The prediction is performed by feeding the utterance embeddings to a simple MLP. In \autoref{tab:ablation}, we report the results on \texttt{MIAM}. Results demonstrate an overall higher accuracy when the pretraining is performed on spoken data. This supports the choice of \texttt{OPS} as pretraining corpora and demonstrates that the origin of the pretraining data matters.

\begin{table*}[]
    \centering
    \begin{tabular}{c|ccccc|c}\hline
                &  \texttt{VM2}  & \texttt{Map Task} & \texttt{Dihana} &  \texttt{Loria} & \texttt{Ilisten}  & Total \\\hline
    $m\mathcal{HT}(\theta_{written})$  & 52.8 & 64.6    & 98.1  &  76.5 & \textbf{74.2}    &  73.2\\
    $m\mathcal{HT}_u(\theta_{spoken})$  & \textbf{53.0}  & \textbf{67.3}   & \textbf{98.3}  & \textbf{78.5} & 74.0    & \textbf{74.2}\\
    \hline\end{tabular}
    \caption{Ablation studies on pretraining data. We report the accuracy on \texttt{MIAM} for the $m\mathcal{HT}$. $m\mathcal{HT}_u(\theta_{spoken})$ stands for the model pretrained with the utterance level loss $m\mathcal{L}^u$ on spoken data and $m\mathcal{HT}(\theta_{written})$ stands for a hierarchical encoder where sentence embeddings is computed using a pretrained BERT encoder.} 
    \label{tab:ablation}
\end{table*}

\end{document}